\documentclass[10pt,twocolumn,letterpaper]{article}

\usepackage{wacv}              

\usepackage{colortbl}

\newcommand{\cmark}{\(\checkmark\)}

\definecolor{oursgreen}{RGB}{219,238,224}

\usepackage{cuted}

\definecolor{wacvblue}{rgb}{0.21,0.49,0.74}
\usepackage[pagebackref,breaklinks,colorlinks,allcolors=wacvblue]{hyperref}

\def\wacvPaperID{1512} 
\def\confName{WACV}
\def\confYear{2027}

\title{MV-Sign: Multi-View 3D Motion Integration and Enhanced Text-to-Motion Conditioning for Sign Language Production}

\author{
Singyu Muk$^{1}$ \enspace
Kaustubh Kundu$^{2}$ \enspace
Kunyuan Xie$^{2}$ \enspace
Weijie Wang$^{3,*}$ \enspace
Kalin Stefanov$^{4,*}$\\
$^{1}$Independent Researcher, Australia \quad
$^{2}$Monash University, Australia\\
$^{3}$University of Trento, Italy \quad
$^{4}$Japan Advanced Institute of Science and Technology, Japan\\
{\tt\small
ronggong000@gmail.com \quad
Kaustubh.Kundu@monash.edu \quad
kunyuan.xie@monash.edu}\\
{\tt\small
weijie.wang@unitn.it \quad
kalins@jaist.ac.jp}
}

\begin{document}

\twocolumn[{%
\renewcommand\twocolumn[1][]{#1}%
\maketitle

\begin{center}
\captionsetup{type=figure}
\includegraphics*[width=\textwidth]{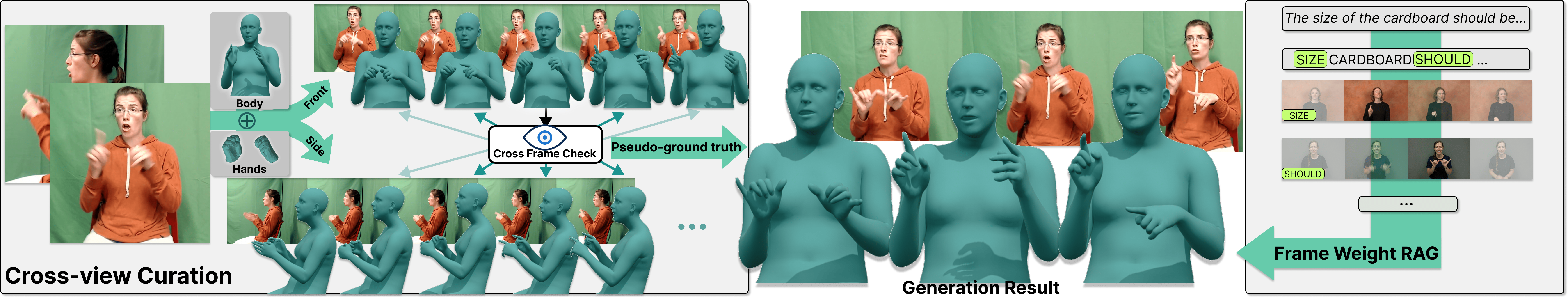}
\captionof{figure}{\textbf{MV-Sign overview.} On the left, our multi-view data curation pipeline reconstructs high-quality 3D signing motion from noisy videos by combining per-view hand fusion, spatial-temporal alignment, and cross-view refinement. On the right, our generative model conditioned on pseudo-gloss and keyframe-aware retrieval prompts produces accurate continuous signing motion from text input.}
\label{fig:teaser}
\end{center}%
}]

\begingroup
\renewcommand{\thefootnote}{}
\footnotetext{* Corresponding author.}
\endgroup

\begin{abstract}
Sign language production (SLP) aims to generate sign language motion from text.
However, current 3D SLP methods suffer from noisy supervision and weak text conditioning.
Relying on pseudo-3D motion lifted from single-view frontal videos introduces severe self-occlusion and loses fine-grained hand details, degrading downstream production.
To address this, we propose a framework that integrates multi-view 3D motion data and optimizes text conditioning.
First, we introduce \textbf{How2Sign-MV3D}, a high-fidelity dataset curated via cross-view fusion to resolve occlusions and filter non-signing segments.
Second, we design a robust SLP model incorporating a sign-aware spatio-temporal VQ-VAE (ST-VQVAE) for joint body-hand tokenization, a pseudo-gloss adapter, and keyframe-aware retrieval to enhance text-to-motion alignment.
Evaluated on the How2Sign benchmark, our framework reduces body and hand DTW-PA-JPE by \textbf{32.8\%} and \textbf{20.9\%} respectively compared to the state-of-the-art.
Finally, we release two more new high-quality 3D datasets, \textbf{ASLCitizen-3D} and \textbf{MM-WLAuslan-MV3D}, alongside \textbf{SGNify-MV} as a multi-view video benchmark to support broader community research.
\end{abstract}
\section{Introduction}
\label{sec:introduction}

Sign languages are the main way of communication in deaf communities~\cite{yin2021including,gong2024llms}.
Sign language production (SLP), also called text-to-sign generation, aims to produce a signing motion from spoken language text.
Unlike sign language translation and understanding, which have recently advanced through large-scale pretraining and multimodal language models~\cite{hu2023signbert+,wong2024sign2gpt,li2025uni,tan2025multilingual}, SLP must produce motion that is both semantically accurate and physically realistic.
These requirements make the availability of high-quality 3D signing motion data a central bottleneck.

Recent SLP methods supervise sign generation by fitting SMPL-X~\cite{pavlakos2019expressive} to frontal-view videos~\cite{duarte2021how2sign, baltatzis2024neural,yu2024signavatars,zuo2025signs} using datasets such as How2Sign~\cite{duarte2021how2sign}.
Although How2Sign contains side-view videos, they suffer from temporal misalignment and heavy occlusions.
Because of this, the side view is ignored by most work.
However, How2Sign front-view videos also have significant issues, including motion blur, low frame rates, editing errors, and self-occlusion.
These issues lead to noisy and incorrect pseudo-3D motion data and are propagated downstream into production.

To address these issues, we propose \textbf{MV-Sign}, a unified framework coupling asymmetric cross-view data curation with a denoised 3D SLP model. Project resources and
dataset release information are available at \url{https://github.com/ronggong000/MV-Sign}. Our main contributions are:
\begin{itemize}
\item We propose an asymmetric cross-view motion curation pipeline that uses side-view evidence to repair occlusions, blur, and tracking failures in frontal videos.
\item We design a 3D SLP model integrating a sign-aware spatio-temporal VQ-VAE (ST-VQVAE) for joint body--hand tokenization, a pseudo-gloss adapter, and keyframe-aware retrieval.
\item We construct and release three 3D sign language datasets (\textbf{How2Sign-MV3D}, \textbf{ASLCitizen-3D}, and \textbf{MM-WLAuslan-MV3D}) alongside \textbf{SGNify-MV}, a controlled multi-view diagnostic benchmark.
\end{itemize}
\section{Related Work}
\label{sec:related_work}

\subsection{3D Signing Data and Motion Curation}
Progress in sign language research has long been constrained by data availability~\cite{bragg2019sign}.
Early benchmarks focus on isolated sign recognition or dictionary-style resources~\cite{athitsos2008american,zahedi2005combination,von2008significance,von2010signum,braffort2010sign,efthimiou2010dicta,efthimiou2012dicta,joze2018ms,li2020word}, while larger corpora such as BSL-1K, BBC-Oxford BSL, the Public DGS Corpus, PHOENIX-2014T, and How2Sign enable sentence-level recognition and generation research~\cite{albanie2020bsl,albanie2021bbc,schembri2013building,hanke2020extending,huang2018video,camgoz2018neural,duarte2021how2sign}.
Although recent multi-view datasets such as MM-WLAuslan~\cite{shen2024mm} provide synchronized videos from multiple views, most sign language resources remain 2D.
Therefore, 3D SLP commonly relies on pseudo-3D motion by fitting parametric models such as SMPL-X~\cite{pavlakos2019expressive}, MANO~\cite{romero2022embodied}, and FLAME~\cite{li2017learning} to videos.
Consequently, SLP pipelines such as NSA~\cite{baltatzis2024neural}, SignAvatars~\cite{yu2024signavatars}, SOKE~\cite{zuo2025signs}, and avatar reconstruction methods~\cite{kundu2026dexavatar} recover sign motion from videos using general whole-body recovery methods~\cite{kocabas2021pare,rong2021frankmocap,feng2021collaborative,zhang2023pymaf,lin2023one,cai2023smpler,lugaresi2019mediapipe,sun2024aios,li2024human}. In parallel, dedicated hand reconstruction methods, including ACR~\cite{yu2023acr}, HaMeR~\cite{pavlakos2023reconstructinghands3dtransformers}, and WiLoR~\cite{potamias2025wilor}, address fine-grained 3D hand estimation.
However, signing requires a finer wrist, handshape, and finger fidelity than generic pose recovery.
Unlike multi-view human reconstruction in controlled settings, sign language corpora often have uncalibrated, occluded, or temporally offset side views.
We therefore use the side view(s) as asymmetric evidence to verify and repair front-view failures, framing our contribution as signing-specific 3D motion curation.

\subsection{Sign Language Production}
Early SLP methods use recurrent or sequence-to-sequence pose generation~\cite{hochreiter1997long,van2017neural}, followed by Transformer-based methods such as Progressive Transformer~\cite{vaswani2017attention,saunders2020progressive}.
Recent research has expanded SLP towards pose-guided video generation, diffusion-based 3D signing, and tokenized motion generation~\cite{shi2024pose,baltatzis2024neural,yu2024signavatars}.
Diffusion-based approaches generate sign pose sequences directly from text~\cite{ho2020denoising,rombach2022high,fang2025signdiff,baltatzis2024neural}, sometimes with anatomically informed backbones.
The tokenizer-language-model approaches first discretize the signing motion with VQ-VAE-style tokenizers and then predict the motion codes autoregressively~\cite{yin2024t2s,dong2024signavatar,yu2024signavatars,zuo2025signs}.
SOKE~\cite{zuo2025signs} is especially relevant because it uses separate token streams for the body, left, and right hand for efficient decoding.
Building on SOKE~\cite{zuo2025signs} and SALAD~\cite{hong2025salad}, we retain the three-stream decoder interface but defer part factorization to the tokenizer bottleneck to preserve body--hand coupling.

\subsection{Gloss, Retrieval, and Evaluation}
Gloss-like sequences can provide motion-aligned intermediate representations for SLP, but gold gloss annotations are expensive and often unavailable~\cite{muller2023considerations}.
Therefore, recent methods explore gloss-free modeling, visual-language pretraining, multilingual translation, LLM-assisted translation, and pseudo-glosses derived from text~\cite{zhou2021spatial,yin2022mlslt,yao2023sign,gong2024llms,wong2024sign2gpt,zhang2025towards}.
Following Zhang \etal~\cite{zhang2025towards}, we use vocabulary-constrained pseudo-glosses as a conditioning interface and apply classifier attribution to filter non-lexical preparation and retraction frames from retrieved sign prompts~\cite{saunders2022signing,zuo2024simplebaselinespokenlanguage,zuo2025signs}.
However, sign clips include preparation, retraction, and idle frames, which can consume the limited prompt budget of an autoregressive motion decoder.
We instead use attribution from an isolated sign recognizer to retain informative tokens.
For evaluation, we follow prior motion-space protocols rather than back-translation, since text metrics derived from separate sign language translation systems can be unreliable for pose-based SLP, especially on challenging datasets such as How2Sign~\cite{jiang2025meaningful,tarres2023sign}.

\section{Cross-View Motion Curation}
\label{sec:denoising_3d_motion}

\begin{figure*}[t]
\centering
\includegraphics[width=\textwidth]{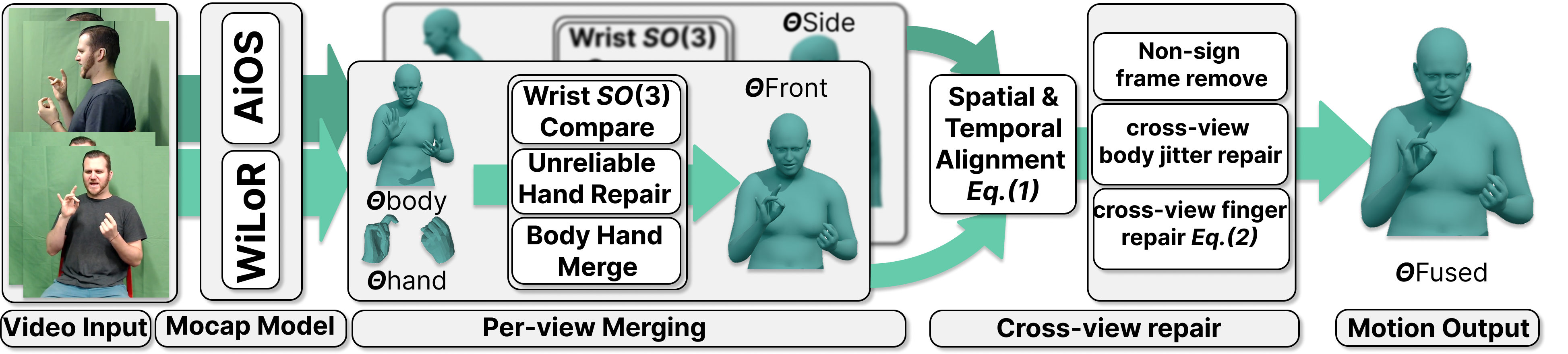}
\caption{\textbf{Cross-view motion curation pipeline.} We first build per-view SMPL-X~\cite{pavlakos2019expressive} sequences using AiOS~\cite{sun2024aios} as a stable body backbone and WiLoR~\cite{potamias2025wilor} as detailed hand evidence. The frontal and side streams are then aligned, synchronized, cleaned of non-signing frames, and used for cross-view body and finger repair. The side view acts as verifier and repair evidence; the final training target for SLP is the refined frontal stream.}
\label{fig:data_pipeline}
\end{figure*}

This section outlines the proposed cross-view 3D data curation pipeline (\Cref{fig:data_pipeline}) used for the creation of \textbf{How2Sign-MV3D}, \textbf{ASLCitizen-3D}, and \textbf{MM-WLAuslan-MV3D}, reconstructed from How2Sign~\cite{duarte2021how2sign}, ASLCitizen~\cite{desai2023aslcitizencommunitysourceddataset}, and MM-WLAuslan~\cite{shen2024mm} datasets, respectively.
Detailed mathematical formulations are deferred to Appendix~\ref{app:motion_curation}.

\subsection{Reliability-Guided Per-View Fusion}

AiOS~\cite{sun2024aios} is used to process multiple frames and reconstruct a stable global reference for the body and wrists.
In parallel, WiLoR~\cite{potamias2025wilor} is used to reconstruct fine handshapes from individual frames.
Since WiLoR is susceptible to failure under motion blur and occlusion, we verify WiLoR against the AiOS wrist references to identify incorrect predictions.
Specifically, a WiLoR hand frame is marked as unreliable if it meets any of three criteria: (i) the hand detection is missing, (ii) the $SO(3)$ wrist orientation severely disagrees with the AiOS reference, or (iii) the wrist exhibits severe angular jitter unsupported by the stable AiOS motion.
Once these unreliable frames are identified, the corrupted wrist and finger intervals are repaired in rotational space using boundary propagation and interpolation across frames.
The final fused per-view SMPL-X~\cite{pavlakos2019expressive} sequence is then obtained by replacing the original AiOS hand parameters with the repaired WiLoR wrist and finger blocks, preserving the stable AiOS global translation, body pose, and shape.
Detailed mathematical formulations are provided in Appendix~\ref{app:sub_per_view_fusion}, while view-specific detection statistics and examples are presented in Appendix~\ref{app:afs6-detection}.

\subsection{Alignment, Synchronization, Non-Sign Frames}

To compare the streams from different cameras, we first extract the 3D joint coordinates $J^v_t$ from the fused per-view SMPL-X sequences obtained in the previous step.
However, the streams cannot be compared directly because the side-view pose estimates are globally shifted, and the videos may have a temporal offset.
To resolve this, we first apply a robust spatial alignment followed by a temporal synchronization search that minimizes the Mean Per Joint Position Error (MPJPE):

\begin{equation}
\delta^\star
=
\arg\min_{\delta\in\mathcal D}
\frac{1}{T_\delta |\mathcal{B}|}
\sum_{t\in\mathcal I_\delta}
\sum_{j \in \mathcal{B}}
\left\|
J^{\mathrm{f}}_{t,j}
-
\tilde J^{\mathrm{s}}_{t-\delta,j}
\right\|_2 .
\end{equation}

Here, $\delta^\star$ is the optimal temporal offset selected from a search range $\mathcal D$ (\ie, $[-8, 8]$ frames), $\mathcal I_\delta$ denotes the set of overlapping frame indices, and $T_\delta=|\mathcal I_\delta|$. A positive $\delta$ aligns source frame $t-\delta$ with target frame $t$.
The notation $J^{\mathrm{f}}_{t,j}$ denotes the 3D position of joint $j$ in the frontal view.
The subset $\mathcal B$ restricts the calculation to the $22$ stable body joints, deliberately excluding the hand joints because they are the noisiest parts before refinement.
The term $\tilde J^{\mathrm{s}}_{t,j}$ represents the side-view joints after a three-stage spatial alignment to the frontal view (which calculates relative frame-wise rotations, extracts a stable average rotation via singular value decomposition (SVD) to eliminate jitter, and aligns the global centroids and pelvis).

Even after synchronization, continuous signing clips can include preparation, ending motion, pauses, and recording artifacts.
Pause and neutral position cues have long been used for continuous sign segmentation~\cite{khan2014pause}, and recent approaches also treat temporal localization as a central problem~\cite{huang2018video,renz2021sign}.
This is consistent with high-quality signing resources that annotate or curate the signing interval rather than treating all recorded frames as lexical motion.
We therefore mask out these non-signing frames using a combination of geometric and motion cues.
Specifically, a frame is classified as active signing only when three conditions are simultaneously met: (i) sufficient mouth activity, (ii) significant finger motion, and (iii) sufficient wrist activity gated by scale-normalized wrist--elbow geometry.
Long inactive runs are removed jointly from all views, while short gaps are bridged via interpolation to avoid discontinuities.
The conjunction of these cues provides a high-precision mask: wrist position alone confuses held rest poses with signs, while finger or mouth motion alone can be triggered by incidental activity.
To ensure optimal tracking accuracy, this activity detection is performed solely on the frontal view, but the resulting temporal mask is applied to truncate and edit all views consistently.
Further details on this step are given in Appendix~\ref{app:sub_sync_and_clean}.

\subsection{Cross-View Body and Finger Repair}

With the streams synchronized and non-signing frames removed, we cross-verify the sequences to repair abrupt failures.
For body joints, we assume that a true fast signing motion should be consistently visible in all cameras.
To quantify this, we compute the frame-to-frame geodesic angular velocity for each joint in $SO(3)$.
A spatial jitter is flagged for correction only if its angular velocity exhibits a severe spike in the target frontal view but remains strictly below a low-jitter threshold in the opposite view.
This mutual-exclusion condition ensures we only target artifactual glitches, preventing the suppression of legitimate, consistent fast movements.
The corrupted frames are then inpainted using spherical linear interpolation and smoothed.

The most sign-critical residual error is missing finger extensions.
Because single-frame estimators struggle to reconstruct fast-moving individual fingers under severe motion blur, we design a targeted repair mechanism based on isolated extension statistics.
First, we extract the $X$-axis flexion component in each joint's geometric finger frame and map it to a normalized straightness score $S_k\in[0,1]$ for each finger $k$ (Appendix~\ref{app:curation_hand_frames}).
We then evaluate a reliable source observation (denoted as $\mathrm{src}$, representing either the opposite view or a temporally adjacent frame) to detect anomalously isolated extensions.
Specifically, a finger $k$ is flagged as confidently extended if its straightness exceeds $1.15$ times the mean of the remaining fingers, while the variance of the remaining fingers is strictly below a threshold $\tau_{\mathrm{var}}$.
This geometric filter effectively isolates intended pointing or gesturing motions from a globally bent hand.
Among the candidates satisfying the above isolation conditions, we select the most prominent one $k^\star = \arg\max_k E^{\mathrm{src}}_{k}$, where $E^{\mathrm{src}}_{k} = S^{\mathrm{src}}_{k} - \mu(S^{\mathrm{src}}_{\setminus k})$ is the excess straightness.
We then verify if this specific extended finger has falsely collapsed in the target destination observation ($\mathrm{dst}$) due to motion blur, and apply a bounded directional repair:

\begin{equation}
\begin{aligned}
c_{k^\star} &= \mathbf{1}\!\left[S^{\mathrm{dst}}_{k^\star}<0.65S^{\mathrm{src}}_{k^\star}\right],\\
\phi^{\mathrm{dst}}_{x,j}
&\leftarrow \operatorname{sign}(\phi^{\mathrm{dst}}_{x,j})
\max\!\left(|\phi^{\mathrm{dst}}_{x,j}|-c_{k^\star}\lambda w_j,0\right).
\end{aligned}
\end{equation}
Here, $c_{k^\star}=1$ triggers the repair if the target straightness falls below $65\%$ of the reliable source observation.
The variable $\phi_{x,j}$ is the adjusted-frame flexion angle of joint $j$ in the target finger, with $w=(0.7,0.7,0.5)$.
The scale $\lambda=1.2E^{\mathrm{src}}_{k^\star}/\sum_{i=1}^{2}a_iw_i/u_{k^\star,i}$ converts normalized straightness to angular units, using $a=(0.7,0.3)$ and the joint bounds $u$ specified in Appendix~\ref{app:curation_finger_repair}.
This two-pass process, first anchoring on the opposite view, then propagating to adjacent frames ensures that we only restore reliably observed extensions lost to blur, without forcing spurious open-hand artifacts.
Further details on this step are given in Appendix~\ref{app:sub_cross_view_repair}.
\section{Conditional Sign Language Production}
\label{sec:denoised_slp_model}

This section outlines the proposed tokenizer-language-model SLP method. 
The method retains the efficient three-stream interface of SOKE~\cite{zuo2025signs} and introduces information connectivity across tokenizer streams and enhanced retrieval.
Detailed mathematical formulations are deferred to Appendix~\ref{app:tokenizer} and Appendix~\ref{app:generation}.

\subsection{Sign-Aware ST-VQVAE}
\label{sec:st_vqvae}

\begin{figure}[t]
\centering
\includegraphics[width=\columnwidth]{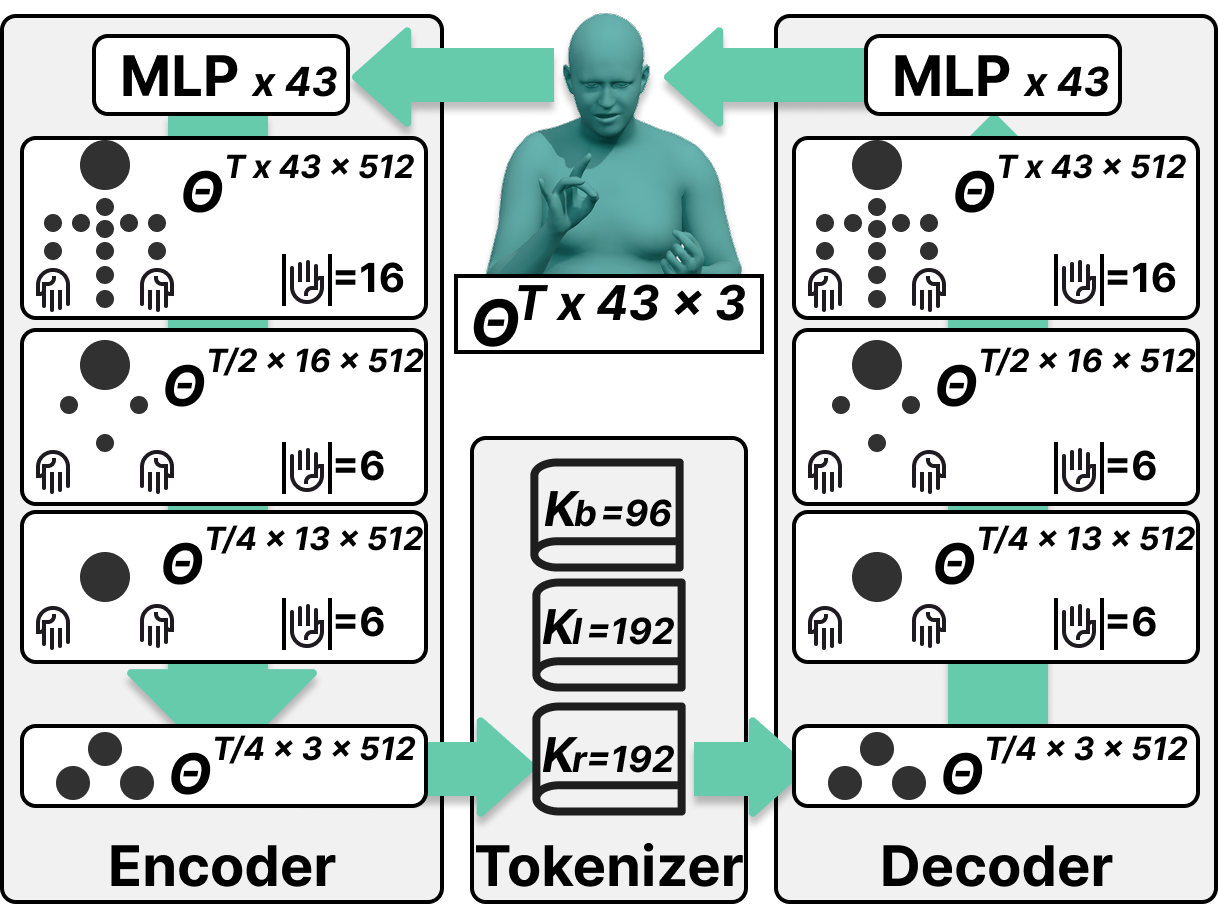}
\caption{\textbf{Sign-aware ST-VQVAE.} Unlike independent body/hand tokenizers, our encoder processes the 43-joint upper-body and hands skeleton jointly, pools it to three semantic slots only at the bottleneck, and quantizes the slots with body, left, and right hand codebooks. This preserves body-hands coupling while retaining the three-token-stream interface used by autoregressive SLP models.}
\label{fig:vqvae_pipeline}
\end{figure}

SOKE discretizes different parts (\ie, body, left, and right hand) with a separate VQ-VAE.
This matches the decoder interface, but removes cross-part interaction inside the tokenizer.
We adapt the skeleton-aware latent hierarchy VAE of SALAD~\cite{hong2025salad} to sign motion: instead of modeling a continuous full-body motion skeleton, our encoder pools a $43$-joint upper-body and hands skeleton into three terminal slots aligned with the decoder interface (\Cref{fig:vqvae_pipeline}).
Sign meaning depends on coordinated body, wrist, and finger articulation: the same handshape can be interpreted differently depending on its trajectory, orientation, and the opposite hand.
We therefore encode the skeleton jointly and factorize only at the bottleneck; each token stream is predicted separately but is formed from a full signing configuration.

As illustrated in Figure~\ref{fig:vqvae_pipeline}, given an SMPL-X motion sequence, our encoder progressively pools the spatial skeleton from $43$ joints down to $3$ terminal slots, while temporally downsampling the sequence to one-quarter of its original frame rate. Spatial resolutions correspond to progressive anatomical groupings. This results in a final continuous latent sequence where each time step contains three $512$-dimensional semantic slots representing the body, left hand, and right hand. These continuous slot representations are then quantized into discrete tokens using part-specific codebooks. Because hands require more capacity to capture rich lexical handshape variations, we allocate larger codebook sizes for the left and right hands ($192$ each) compared to the body codebook ($96$). The decoder mirrors the encoder and reconstructs the SMPL-X motion.
We train the tokenizer with rotation reconstruction, mesh-level geometric supervision, and vector-quantization regularization:

\begin{equation}
\mathcal L_{\mathrm{tok}}
=
\lambda_{\mathrm{rot}}\mathcal L_{\mathrm{rot}}
+
\lambda_{\mathrm{mesh}}\mathcal L_{\mathrm{mesh}}
+
\lambda_{\mathrm{vq}}\mathcal L_{\mathrm{vq}}.
\end{equation}
Rotation supervision preserves local articulation, while mesh-level supervision penalizes visible geometric errors after SMPL-X decoding.
This combination matters for sign motion because a small rotation error can still create a visibly wrong fingertip or wrist trajectory.

\subsection{Pseudo-Gloss Conditioning}
\label{sec:pseudo_gloss_conditioning}

Raw text is a noisy source interface for SLP because sign languages differ in lexical choice, word order, and alignment granularity.
It is also a weak interface for retrieval: an isolated sign dictionary is indexed by sign-level labels (glosses) rather than text tokens.
Therefore, we convert text sentences \(x\) into a pseudo-gloss sequence \(g=\Phi_{\mathrm{LLM}}(x)\) offline using GPT-5.2 with few-shot prompting.
The prompt follows the appendix template of Zhang \etal~\cite{zhang2025towards}, with an additional ASLCitizen~\cite{desai2023aslcitizencommunitysourceddataset} vocabulary constraint.
We use pseudo-gloss as a practical intermediate representation; the goal is to provide the generator and retrieval module with a source sequence closer to sign-level motion.
To maintain SOKE~\cite{zuo2025signs} generation structure, we introduce a lightweight gloss adapter, consisting of a word-level gloss tokenizer, $512$-d learned gloss and positional embeddings, a $4$-layer Transformer encoder ($8$ heads, $2048$ FFN, $0.1$ dropout), and a linear projection to the $1024$-d mBART hidden space.
The downstream decoder remains unchanged.
Pseudo-gloss evaluation settings, reference-agreement results, and vocabulary coverage are provided in Appendix~\ref{app:pseudogloss_validation_v3}.
\subsection{Keyframe-Aware Retrieval}
\label{sec:keyframe_retrieval}

\begin{figure}[t]
\centering
\includegraphics[width=\columnwidth]{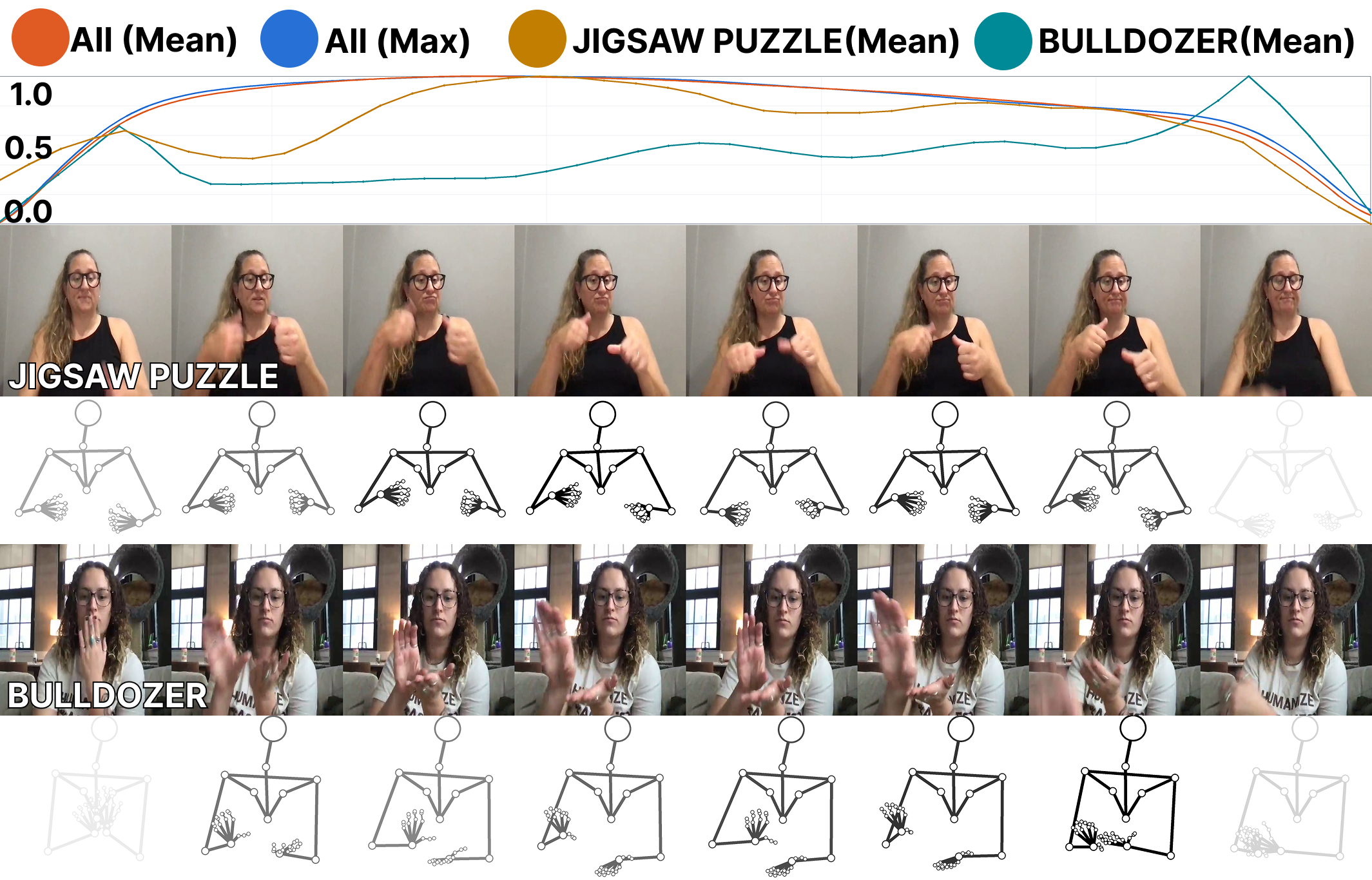}
\caption{\textbf{Keyframe distribution.} In the ASLCitizen~\cite{desai2023aslcitizencommunitysourceddataset} dataset, the average and maximum IG~\cite{sundararajan2017axiomaticattributiondeepnetworks} saliency of joints over time follow a rough Gaussian trend overall yet vary widely across signs, motivating per-clip rather than global-envelope selection.}
\label{fig:saliency_retrieval}
\end{figure}

In SOKE-style retrieval, isolated sign clips are processed into discrete tokens via a VQ-VAE to serve as auxiliary prompts.
However, these clips often include non-lexical preparation and retraction motions; these frames can be treated as retrieval prompt noise.
We identify lexically informative regions in isolated sign clips using a TCN-based~\cite{bai2018empiricalevaluationgenericconvolutional} classifier trained on ASLCitizen~\cite{desai2023aslcitizencommunitysourceddataset} (which provides approximately 20 video instances per sign).
Because extracting attribution maps from the same data used for training yields overfitted and biased saliency, we adopt a $3$-fold cross-validation protocol: we train identical TCNs on two data splits and extract inferences on the held-out third.
We compute these out-of-fold attributions using Integrated Gradients (IG)~\cite{sundararajan2017axiomaticattributiondeepnetworks}, empirically choosing it over alternatives such as Grad-CAM~\cite{Selvaraju_2019}.
(See Appendix~\ref{app:token_aligned_window} for IG vs.\ Grad-CAM comparisons and soft-weighting ablations.)
For each isolated sign clip, the frozen ST-VQVAE encodes the motion into discrete tokens, the TCN classifier processes the raw $43$-joint skeleton and produces a joint-time attribution map.
To align this continuous saliency with the discrete generator, we first average the IG values across all $43$ joints and normalize the scores within each clip from $0$ to $1$ (\Cref{fig:saliency_retrieval}).
Since the main SLP generator operates at a $4\times$ temporally downsampled rate, we apply a corresponding $4\times$ temporal average pooling to the 1D saliency array to match the downsampled token timeline.
Finally, a constrained retrieval budget of up to $10$ tokens is selected from the highest-weighted regions of the encoded sequence and concatenated with the source condition.
This processing strategy keeps the retrieval module compatible with the three-stream token interface, preventing the concatenation of overly long input sequences from degrading the generator's training, while ensuring the prompt budget is strictly concentrated on informative lexical motion.

\subsection{Autoregressive Token Prediction}
\label{sec:autoregressive_prediction}

\begin{figure}[t]
\centering
\includegraphics[
width=0.98\columnwidth,
height=0.28\textheight,
keepaspectratio
]{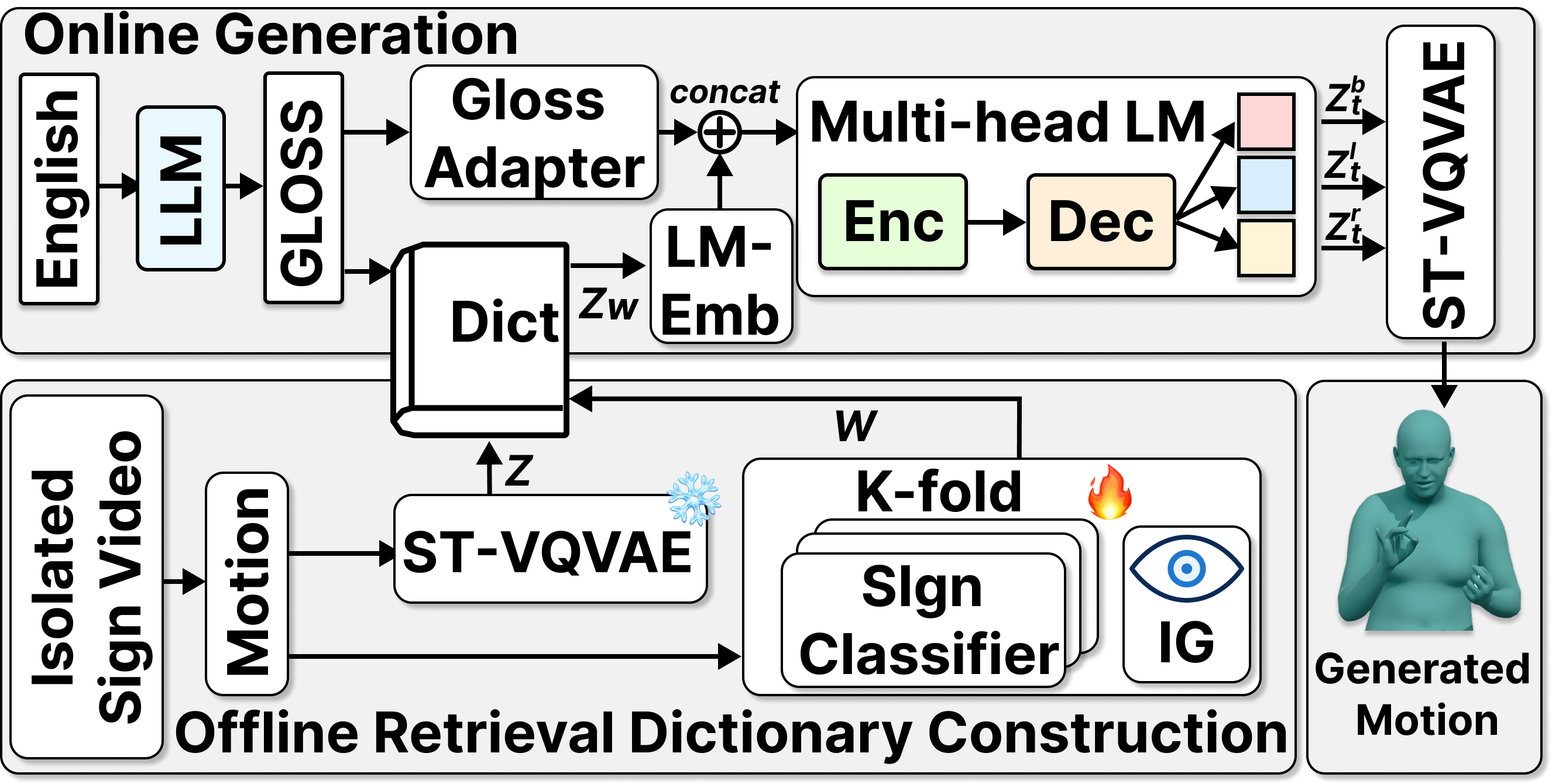}
\caption{\textbf{SLP with sign classifier enhanced retrieval.} We train an auxiliary isolated-sign classifier and extract joint-time saliency $W$ using Integrated Gradients (IG). After temporal pooling to the ST-VQVAE token rate, we use the saliency scores to select a fixed 10-token window from the encoded isolated-sign sequence $Z_w$. The selected tokens serve as a compact retrieval prompt for autoregressive sign generation.}
\label{fig:generation_pipeline}
\end{figure}

The final source representation concatenates the gloss-adapted sequence and retrieval prompt embeddings:

\begin{equation}
H_{\mathrm{src}}=[\hat H_g;H_{\mathrm{ret}}],
\end{equation}
with \(H_{\mathrm{src}}=\hat H_g\) when no retrieval entry is available.
The generator predicts the three token streams autoregressively:

\begin{equation}
\mathcal L_{\mathrm{gen}}
=
-\sum_{t=1}^{T'}
\sum_{p\in\{b,l,r\}}
\log p_\theta
\left(
z_t^{(p)}
\mid
H_{\mathrm{src}}, Z_{<t}
\right).
\end{equation}
Training uses teacher-forced cross-entropy over the three vocabularies.
At inference time, the predicted tokens are decoded by the frozen ST-VQVAE decoder into continuous SMPL-X motion (\Cref{fig:generation_pipeline}).
\section{Experiments}
\label{sec:experiments}

We evaluate the \textbf{MV-Sign} framework for data and generation quality using the proposed \textbf{How2Sign-MV3D} dataset.

\subsection{Experimental Setup}
\label{sec:experimental_setup}

\noindent \textbf{Datasets.}
The regenerated How2Sign-MV3D release follows the How2Sign train/validation/test partition~\cite{duarte2021how2sign} and contains $30{,}903$, $1{,}711$, and $2{,}325$ completed motion sequences, respectively. Of the $35{,}241$ clips in our processing manifest, $289$ were excluded by input screening and $13$ failed processing. Clips use the official manually realigned sentence timestamps; motion-side non-sign removal reduces untranscribed waiting intervals without additional text reannotation (Appendix~\ref{app:afs6-text-motion}).
The tokenizer and generator are trained only on \textbf{How2Sign-MV3D}.
ASLCitizen~\cite{desai2023aslcitizencommunitysourceddataset} is used only for isolated sign retrieval and saliency estimation.
\textbf{SGNify-MV} contains $57$ isolated signs with ground-truth meshes and is used in the Appendix as a diagnostic rather than as the main evidence for cross-view curation on continuous signing.

\noindent \textbf{Metrics.}
For SLP on How2Sign, we follow prior work and report DTW-JPE and DTW-PA-JPE for body and hand regions~\cite{baltatzis2024neural,zuo2025signs}.
DTW aligns generated and reference sequences in time; PA denotes Procrustes alignment before computing joint error.
For tokenizer reconstruction, the input and reconstruction are already frame-aligned, so Table~\ref{tab:tokenizer_reconstruction} and Table~\ref{tab:tokenizer_loss_ablation} report PA-JPE for joints and PA-VPE for vertices without DTW.
PA is computed on the evaluated region: body uses the upper-body joint set, and hand scores align the left and right hands separately before averaging.
Prior 3D SLP methods for How2Sign, including NSA~\cite{baltatzis2024neural} and SOKE~\cite{zuo2025signs}, build pseudo-3D training targets and then evaluate held-out text-to-motion generation against the corresponding How2Sign motion-space protocol.
We therefore treat Table~\ref{tab:sota_how2sign} as an end-to-end pipeline comparison, where data curation and generator design both affect the final test set error, rather than as a pure model comparison under fixed ground-truth 3D labels.
The \textbf{SGNify-MV} diagnostic in the Appendix reports translation-removed vertex-to-vertex error (TR-V2V, mm) on the upper body excluding the face (UBody-F), left, and right hand.
We omit back-translation because recent pose-based SLP evaluation studies note that such metrics require dataset- or direction-specific pose-to-text evaluators whose code and weights are rarely standardized or released, making private BLEU scores difficult to compare~\cite{jiang2025meaningful,tarres2023sign}.
We therefore focus on motion-space metrics and qualitative motion inspection.

\noindent \textbf{Implementation details.}
The ST-VQVAE is trained for $150$ epochs with AdamW, learning rate \(2\times10^{-4}\), batch size $32$, latent width \(d=512\), temporal downsampling factor \(4\), and codebook sizes \((96,192,192)\).
The autoregressive generator is an mBART-large SOKE-style backbone with three token heads and is trained with teacher forcing.
Pseudo-glosses are generated offline with GPT-5.2 using the few-shot prompt style of Zhang \etal~\cite{zhang2025towards}, with an added ASLCitizen vocabulary constraint.
Retrieval uses $10$-token prompts selected from isolated sign token sequences.
Inference uses greedy decoding for deterministic comparison.

\subsection{How2Sign-MV3D Results}
\label{sec:data_quality_eval}

\begin{table}[t]
\centering
\caption{\textbf{Quantitative How2Sign-MV3D results.} We report DTW-PA-JPE only. Reported rows are taken from prior work and hand values are averages}
\label{tab:data_quality_downstream}
\small
\setlength{\tabcolsep}{1pt}
\begin{tabular}{lccc}
\toprule
\textbf{Training} & \textbf{UBody-F $\downarrow$} & \textbf{LHand $\downarrow$} & \textbf{RHand $\downarrow$} \\
\midrule
SOKE (3 datasets, reported) & 6.82 & \multicolumn{2}{c}{2.35 (hands avg.)} \\
SOKE (NSA's H2S, reported) & 7.92 & \multicolumn{2}{c}{3.07 (hands avg.)} \\
\midrule
SOKE (\textbf{H2S-MV3D}-only) & \textbf{4.58} & \textbf{1.92} & \textbf{1.98} \\
\bottomrule
\end{tabular}
\end{table}

To test whether denoised motion improves the actual SLP training signal, we compare SOKE-style training and evaluation under different data conditions.
Table~\ref{tab:data_quality_downstream} provides downstream evidence that training on \textbf{How2Sign-MV3D} yields lower motion-space errors within a SOKE-style pipeline, reducing both upper-body and hand errors relative to the reported SOKE results.

\subsection{SLP Results}
\label{sec:slp_comparison}

\begin{table}[t]
\centering
\caption{\textbf{Quantitative SLP results.} Following prior SLP work, we report DTW-PA-JPE and DTW-JPE for body and hands. Since each method relies on pseudo-3D supervision, this table compares complete pipelines under prior motion-space protocols, not isolated generator architectures under shared ground-truth. Baseline rows follow the SOKE comparison.}
\label{tab:sota_how2sign}
\small
\begin{tabular}{lcccc}
\toprule
\textbf{Method}
& \multicolumn{2}{c}{\textbf{DTW-PA-JPE $\downarrow$}}
& \multicolumn{2}{c}{\textbf{DTW-JPE $\downarrow$}} \\
\cmidrule(lr){2-3}\cmidrule(lr){4-5}
& \textbf{Body} & \textbf{Hands} & \textbf{Body} & \textbf{Hands} \\
\midrule
NAR~\cite{hwang2021non} & 13.94 & 11.80 & -- & -- \\
Prog. Trans.~\cite{saunders2020progressive} & 14.15 & 11.57 & 14.74 & 30.17 \\
Text2Mesh~\cite{saunders2022signing} & 13.99 & 13.47 & 15.50 & 32.97 \\
Adv. Train.~\cite{saunders2020adversarial} & 13.78 & 11.17 & -- & -- \\
T2S-GPT~\cite{yin2024t2s} & 11.48 & 6.39 & 12.65 & 18.44 \\
NSA~\cite{baltatzis2024neural} & 7.83 & 7.33 & -- & -- \\
S-MotionGPT~\cite{jiang2023motiongpt} & 11.23 & 4.39 & 12.41 & 13.74 \\
SOKE~\cite{zuo2025signs} & 6.82 & 2.35 & 7.75 & 10.08 \\
\midrule
\textbf{Ours}
& \textbf{4.58}
& \textbf{1.86}
& \textbf{4.98}
& \textbf{7.95} \\
\bottomrule
\end{tabular}
\end{table}

Table~\ref{tab:sota_how2sign} compares our framework with prior text-to-sign and 3D SLP methods under the How2Sign motion-space protocols used in this line of work.
Because How2Sign has no ground-truth 3D motion, these systems evaluate complete pipelines: each method first constructs or adopts pseudo-3D supervision, then trains a generator and evaluates held-out test samples in motion space.
Our framework obtains the lowest body and hand errors among the compared pipelines, and the qualitative comparisons remain important for interpreting differences caused by pseudo-target noise.

\subsection{Ablations}
\label{sec:ablations}

\begin{table}[t]
\centering
\caption{\textbf{Tokenizer ablation.} Using the same codebook size and dimension, our tokenizer reaches better reconstruction results compared to the SOKE tokenizer. DTW is not used because tokenizer reconstruction is frame-aligned.}
\label{tab:tokenizer_reconstruction}
\small
\setlength{\tabcolsep}{3pt}
\begin{tabular}{lcccc}
\toprule
\textbf{Method}
& \multicolumn{2}{c}{\textbf{PA-JPE $\downarrow$}}
& \multicolumn{2}{c}{\textbf{PA-VPE $\downarrow$}} \\
\cmidrule(lr){2-3}\cmidrule(lr){4-5}
& \textbf{Body} & \textbf{Hands} & \textbf{All} & \textbf{Hands} \\
\midrule
SOKE (VQ-VAE) & 19.37 & 6.65 & -- & -- \\
SOKE (VQ-VAE, H2S-MV3D) & \textbf{14.00} & 6.24 & 18.12 & 6.07 \\
\midrule
\textbf{Ours (ST-VQVAE)} & 14.21 & \textbf{5.84} & \textbf{16.59} & \textbf{5.84} \\
\bottomrule
\end{tabular}
\end{table}
\begin{table}[t]
\centering
\caption{\textbf{Tokenizer loss ablation.} Mesh supervision improves global geometry, while the hand-weighted combined loss gives the best hand reconstruction. DTW is not used because tokenizer reconstruction is frame-aligned.}
\label{tab:tokenizer_loss_ablation}
\small
\begin{tabular}{lcccc}
\toprule
\textbf{Loss}
& \multicolumn{2}{c}{\textbf{PA-JPE $\downarrow$}}
& \multicolumn{2}{c}{\textbf{PA-VPE $\downarrow$}} \\
\cmidrule(lr){2-3}\cmidrule(lr){4-5}
& \textbf{Body} & \textbf{Hands} & \textbf{All} & \textbf{Hands} \\
\midrule
Rotation-only & 18.48 & 6.23 & 21.93 & 6.04 \\
Mesh-only & \textbf{11.61} & 10.24 & \textbf{13.33} & 9.96 \\
Rotation + mesh & 14.21 & \textbf{5.84} & 16.59 & \textbf{5.84} \\
\bottomrule
\end{tabular}
\end{table}
\begin{table}[t]
\centering
\caption{\textbf{Component ablation.} DTW-PA-JPE for the main generation variants and relative gains over the reported SOKE body and hands average. Components provide complementary gains rather than a strictly monotonic improvement in every region. Percentage gains are computed from the displayed scores.}
\label{tab:generation_ablation}
\tiny
\setlength{\tabcolsep}{1pt}
\begin{tabular}{lcccccccccc}
\toprule
\textbf{Setting}
& \textbf{MV3D}
& \textbf{Gloss}
& \textbf{Adapter}
& \textbf{Keyframe}
& \textbf{ST-VQVAE}
& \multicolumn{3}{c}{\textbf{DTW-PA-JPE $\downarrow$}}
& \multicolumn{2}{c}{\textbf{Gain $\uparrow$}} \\
\cmidrule(lr){7-9}\cmidrule(l){10-11}
& & & & & & \textbf{Body} & \textbf{LHand} & \textbf{RHand} & \textbf{Body} & \textbf{Hands} \\
\midrule
SOKE (reported) & -- & -- & -- & Original & Original & 6.82 & \multicolumn{2}{c}{2.35 avg.} & -- & -- \\
\midrule
SOKE (H2S-MV3D) & \cmark & -- & -- & -- & -- & 4.58 & 1.92 & 1.98 & 32.8\% & 17.0\% \\
+ adapter & \cmark & \cmark & \cmark & -- & -- & \textbf{4.49} & 1.93 & 1.99 & \textbf{34.2\%} & 16.6\% \\
+ retrieval & \cmark & \cmark & \cmark & \cmark & -- & 4.52 & 1.91 & 1.97 & 33.7\% & 17.4\% \\
+ ST-VQVAE & \cmark & \cmark & \cmark & \cmark & \cmark & 4.58 & \textbf{1.81} & \textbf{1.91} & 32.8\% & \textbf{20.9\%} \\
\bottomrule
\end{tabular}
\end{table}

Table~\ref{tab:tokenizer_reconstruction} reports that our ST-VQVAE improves hands and mesh reconstruction while retaining the three-stream interface required by the autoregressive decoder.
The result supports the tokenizer design: prediction remains part-wise, but each token is produced after jointly observing the $43$-joint skeleton.
Table~\ref{tab:tokenizer_loss_ablation} reports the tradeoff behind the final objective.
Mesh-only training reduces global vertex and body-joint errors because vertex-space losses are dominated by larger body motion and surface displacement, but this hurts the finer hand reconstruction needed for signing.
We therefore use a combined rotation--mesh objective with extra hand weighting, accepting a modest increase in body/global reconstruction error in exchange for better hand fidelity.

Table~\ref{tab:generation_ablation} reports cumulative model components on \textbf{How2Sign-MV3D}.
Using the reported SOKE body score of \(6.82\) and hands average of \(2.35\) as reference, \textbf{How2Sign-MV3D}, pseudo-gloss adaptation, keyframe retrieval, and the final ST-VQVAE tokenizer improve body DTW-PA-JPE by \(32.8\%\), \(34.2\%\), \(33.7\%\), and \(32.8\%\), respectively, and hands-average DTW-PA-JPE by \(17.0\%\), \(16.6\%\), \(17.4\%\), and \(20.9\%\).
The components do not improve every region monotonically, but they provide complementary benefits: pseudo-gloss improves body alignment, keyframe retrieval concentrates the prompt budget, and ST-VQVAE improves hand-aligned metrics.

\begin{figure*}[t]
\centering
\includegraphics[width=\textwidth]{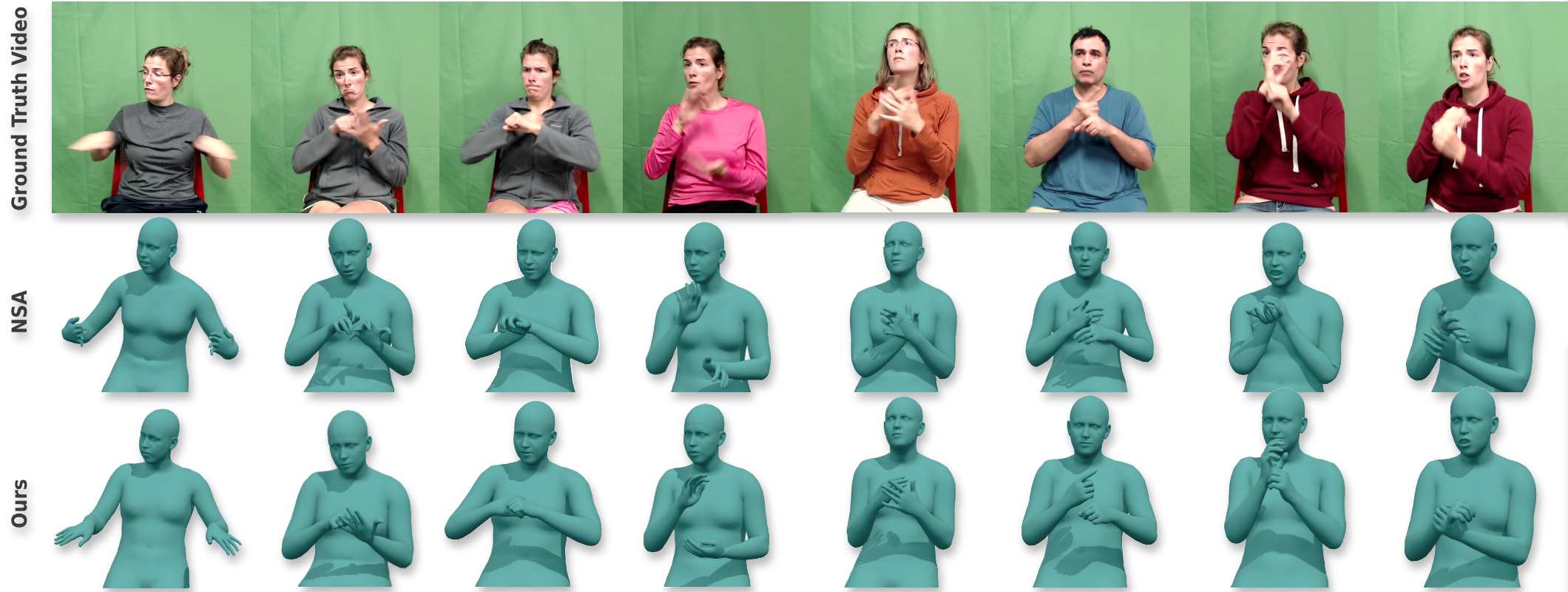}
\caption{\textbf{Qualitative How2Sign-MV3D results.} The figure compares noisy front-view 3D motions with our cross-view-curated motions, emphasizing wrist and finger repair.}
\label{fig:data_quality_compare}
\end{figure*}
\begin{figure*}[t]
    \centering
    \includegraphics[
        width=\textwidth,
        height=0.35\textheight,
        keepaspectratio
    ]{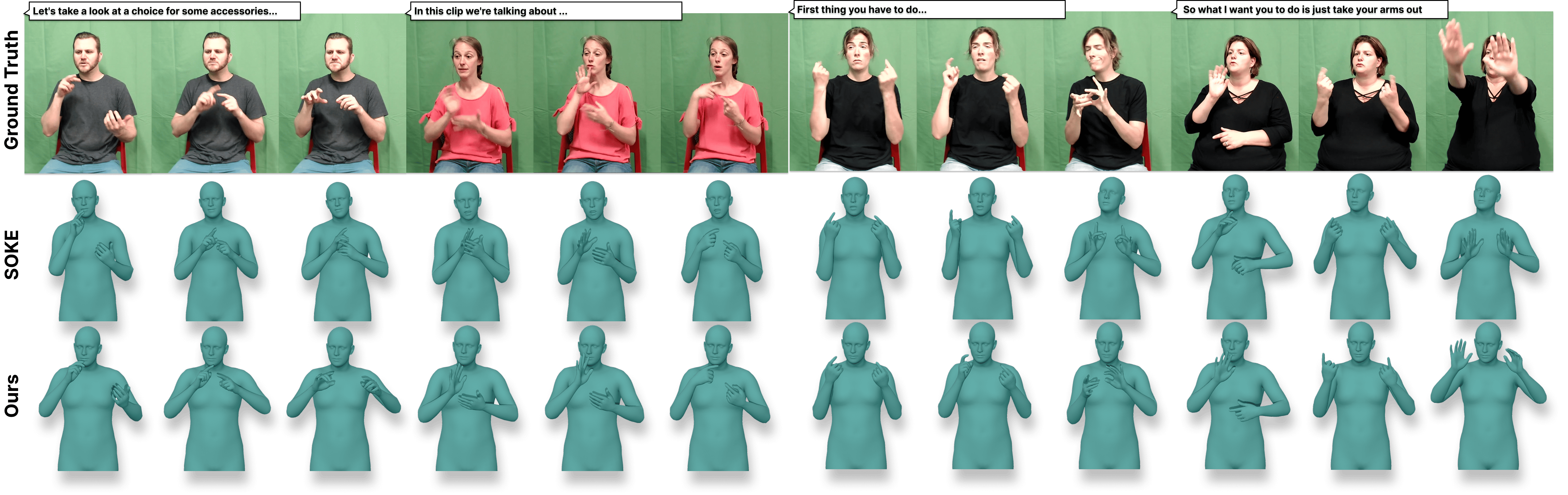}
    \caption{\textbf{Qualitative SLP results.} Comparisons of generated sign sequences from our method (trained with our \textbf{How2Sign-MV3D} pseudo-3D motion) and the state-of-the-art SOKE~\cite{zuo2025signs} (trained with NSA~\cite{baltatzis2024neural} pseudo-3D motion) on How2Sign~\cite{duarte2021how2sign} test set.} 
    \label{fig:generated}
\end{figure*}

\subsection{Qualitative and Diagnostic Analysis}
\label{sec:qualitative_analysis}

Figure~\ref{fig:data_quality_compare} presents qualitative evidence for the data curation stage.
The comparison focuses on wrist and finger artifacts in noisy frontal-view pseudo-3D motions and shows how the refined \textbf{How2Sign-MV3D} target removes view-specific failures that would otherwise propagate into production.
This visual evidence is the main qualitative support for cross-view repair of How2Sign clips; \textbf{SGNify-MV} is kept as a controlled diagnostic in the Appendix.
Figure~\ref{fig:generated} provides representative qualitative SLP results.

The component ablation in Table~\ref{tab:generation_ablation} should be read as a diagnostic of complementary denoising effects rather than as a strictly monotonic component stack.
Pseudo-gloss mainly improves body/source alignment by giving the generator and retrieval module a sign-level conditioning sequence.
Keyframe-aware retrieval slightly improves hands by spending the fixed prompt budget on lexical motion instead of preparation or retraction.
The ST-VQVAE gives the clearest hands gain because it reconstructs body and hands jointly before factorizing into the three token streams used by the decoder.

\section{Conclusion}
\label{sec:conclusion}

We presented \textbf{MV-Sign}, a denoising framework for 3D sign language production.
On the data side, the framework uses cross-view evidence to repair noisy monocular 3D sign motion.
On the model side, the proposed ST-VQVAE, pseudo-gloss adapter, and keyframe-aware retrieval reduce representation, language-conditioning, and retrieval-prompt noise.
How2Sign experiments provide downstream evidence that curated 3D supervision improves SOKE-style training, and the full framework achieves the strongest motion-space metrics among compared SLP pipelines.

\paragraph{Limitations and impact.}
Cross-view evidence is most useful when the frontal view is ambiguous or occluded.
If the frontal view is already clean and the side view is temporally mismatched, side evidence can introduce small noise; our pipeline therefore treats side views as reliability-weighted repair evidence rather than mandatory targets.
The retrieval module estimates joint-time saliency, but the current three-stream token generator consumes an averaged signal; future joint-aware token generators could use the full saliency map.
SLP systems should be developed as assistive tools and not as replacements for human interpreters.

\newpage
{
\small
\bibliographystyle{ieeenat_fullname}
\bibliography{main}

@inproceedings{vaswani2017attention,
  title={Attention Is All You Need},
  author={Vaswani, Ashish and Shazeer, Noam and Parmar, Niki and Uszkoreit, Jakob and Jones, Llion and Gomez, Aidan N and Kaiser, {\L}ukasz and Polosukhin, Illia},
  booktitle={Advances in Neural Information Processing Systems},
  year={2017}
}

@article{li2024human,
  title={Human Mesh Recovery from Arbitrary Multi-view Images},
  author={Li, Xiaoben and Meng, Mancheng and Wu, Ziyan and Chen, Terrence and Yang, Fan and Shen, Dinggang},
  journal={arXiv preprint arXiv:2403.12434},
  year={2024}
}

@inproceedings{gong2024llms,
  title={Llms are good sign language translators},
  author={Gong, Jia and Foo, Lin Geng and He, Yixuan and Rahmani, Hossein and Liu, Jun},
  booktitle={Proceedings of the IEEE/CVF conference on computer vision and pattern recognition},
  pages={18362--18372},
  year={2024}
}

@inproceedings{yin2021including,
  title={Including signed languages in natural language processing},
  author={Yin, Kayo and Moryossef, Amit and Hochgesang, Julie and Goldberg, Yoav and Alikhani, Malihe},
  booktitle={Proceedings of the 59th Annual Meeting of the Association for Computational Linguistics and the 11th International Joint Conference on Natural Language Processing (Volume 1: Long Papers)},
  pages={7347--7360},
  year={2021}
}

@article{hu2023signbert+,
  title={Signbert+: Hand-model-aware self-supervised pre-training for sign language understanding},
  author={Hu, Hezhen and Zhao, Weichao and Zhou, Wengang and Li, Houqiang},
  journal={IEEE Transactions on Pattern Analysis and Machine Intelligence},
  volume={45},
  number={9},
  pages={11221--11239},
  year={2023},
  publisher={IEEE}
}

@misc{hu2024expressivegaussian,
  title={Expressive Gaussian Human Avatars from Monocular RGB Video},
  author={Hu, Hezhen and Fan, Zhiwen and Wu, Tianhao and Xi, Yihan and Lee, Seoyoung and Pavlakos, Georgios and Wang, Zhangyang},
  year={2024},
  eprint={2407.03204},
  archivePrefix={arXiv},
  primaryClass={cs.CV},
  url={https://arxiv.org/abs/2407.03204}
}

@article{li2025uni,
  title={Uni-sign: Toward unified sign language understanding at scale},
  author={Li, Zecheng and Zhou, Wengang and Zhao, Weichao and Wu, Kepeng and Hu, Hezhen and Li, Houqiang},
  journal={arXiv preprint arXiv:2501.15187},
  year={2025}
}

@inproceedings{tan2025multilingual,
  title={Multilingual gloss-free sign language translation: Towards building a sign language foundation model},
  author={Tan, Sihan and Miyazaki, Taro and Nakadai, Kazuhiro},
  booktitle={Proceedings of the 63rd Annual Meeting of the Association for Computational Linguistics (Volume 2: Short Papers)},
  pages={553--561},
  year={2025}
}

@article{wong2024sign2gpt,
  title={Sign2GPT: Leveraging large language models for gloss-free sign language translation},
  author={Wong, Ryan and Camgoz, Necati Cihan and Bowden, Richard},
  journal={arXiv preprint arXiv:2405.04164},
  year={2024}
}

@inproceedings{yao2023sign,
  title={Sign language translation with iterative prototype},
  author={Yao, Huijie and Zhou, Wengang and Feng, Hao and Hu, Hezhen and Zhou, Hao and Li, Houqiang},
  booktitle={Proceedings of the IEEE/CVF International Conference on Computer Vision},
  pages={15592--15601},
  year={2023}
}

@inproceedings{yin2022mlslt,
  title={Mlslt: Towards multilingual sign language translation},
  author={Yin, Aoxiong and Zhao, Zhou and Jin, Weike and Zhang, Meng and Zeng, Xingshan and He, Xiaofei},
  booktitle={Proceedings of the IEEE/CVF conference on computer vision and pattern recognition},
  pages={5109--5119},
  year={2022}
}

@article{zhou2021spatial,
  title={Spatial-temporal multi-cue network for sign language recognition and translation},
  author={Zhou, Hao and Zhou, Wengang and Zhou, Yun and Li, Houqiang},
  journal={IEEE transactions on multimedia},
  volume={24},
  pages={768--779},
  year={2021},
  publisher={IEEE}
}

@inproceedings{baltatzis2024neural,
  title={Neural sign actors: A diffusion model for 3d sign language production from text},
  author={Baltatzis, Vasileios and Potamias, Rolandos Alexandros and Ververas, Evangelos and Sun, Guanxiong and Deng, Jiankang and Zafeiriou, Stefanos},
  booktitle={Proceedings of the IEEE/CVF Conference on Computer Vision and Pattern Recognition},
  pages={1985--1995},
  year={2024}
}

@inproceedings{saunders2020progressive,
  title={Progressive transformers for end-to-end sign language production},
  author={Saunders, Ben and Camgoz, Necati Cihan and Bowden, Richard},
  booktitle={European Conference on Computer Vision},
  pages={687--705},
  year={2020},
  organization={Springer}
}

@inproceedings{shi2024pose,
  title={Pose-guided fine-grained sign language video generation},
  author={Shi, Tongkai and Hu, Lianyu and Shang, Fanhua and Feng, Jichao and Liu, Peidong and Feng, Wei},
  booktitle={European Conference on Computer Vision},
  pages={392--409},
  year={2024},
  organization={Springer}
}

@inproceedings{yu2024signavatars,
  title={Signavatars: A large-scale 3d sign language holistic motion dataset and benchmark},
  author={Yu, Zhengdi and Huang, Shaoli and Cheng, Yongkang and Birdal, Tolga},
  booktitle={European Conference on Computer Vision},
  pages={1--19},
  year={2024},
  organization={Springer}
}

@inproceedings{albanie2020bsl,
  title={BSL-1K: Scaling up co-articulated sign language recognition using mouthing cues},
  author={Albanie, Samuel and Varol, G{\"u}l and Momeni, Liliane and Afouras, Triantafyllos and Chung, Joon Son and Fox, Neil and Zisserman, Andrew},
  booktitle={European conference on computer vision},
  pages={35--53},
  year={2020},
  organization={Springer}
}

@article{albanie2021bbc,
  title={Bbc-oxford british sign language dataset},
  author={Albanie, Samuel and Varol, G{\"u}l and Momeni, Liliane and Bull, Hannah and Afouras, Triantafyllos and Chowdhury, Himel and Fox, Neil and Woll, Bencie and Cooper, Rob and McParland, Andrew and others},
  journal={arXiv preprint arXiv:2111.03635},
  year={2021}
}

@inproceedings{camgoz2018neural,
  title={Neural sign language translation},
  author={Camgoz, Necati Cihan and Hadfield, Simon and Koller, Oscar and Ney, Hermann and Bowden, Richard},
  booktitle={Proceedings of the IEEE conference on computer vision and pattern recognition},
  pages={7784--7793},
  year={2018}
}

@inproceedings{duarte2021how2sign,
  title={How2sign: a large-scale multimodal dataset for continuous american sign language},
  author={Duarte, Amanda and Palaskar, Shruti and Ventura, Lucas and Ghadiyaram, Deepti and DeHaan, Kenneth and Metze, Florian and Torres, Jordi and Giro-i-Nieto, Xavier},
  booktitle={Proceedings of the IEEE/CVF conference on computer vision and pattern recognition},
  pages={2735--2744},
  year={2021}
}

@inproceedings{hanke2020extending,
  title={Extending the Public DGS Corpus in size and depth},
  author={Hanke, Thomas and Schulder, Marc and Konrad, Reiner and Jahn, Elena},
  booktitle={Proceedings of the LREC2020 9th workshop on the representation and processing of sign languages: Sign language resources in the service of the language community, technological challenges and application perspectives},
  pages={75--82},
  year={2020}
}

@inproceedings{huang2018video,
  title={Video-based sign language recognition without temporal segmentation},
  author={Huang, Jie and Zhou, Wengang and Zhang, Qilin and Li, Houqiang and Li, Weiping},
  booktitle={Proceedings of the AAAI conference on artificial intelligence},
  volume={32},
  number={1},
  year={2018}
}

@inproceedings{renz2021sign,
  title={Sign Segmentation with Changepoint-Modulated Pseudo-Labelling},
  author={Renz, Katrin and Stache, Nicolaj C. and Fox, Neil and Varol, G{\"u}l and Albanie, Samuel},
  booktitle={Proceedings of the IEEE/CVF Conference on Computer Vision and Pattern Recognition Workshops},
  pages={3403--3412},
  year={2021}
}

@article{khan2014pause,
  author={Khan, Shujjat and Bailey, Donald G. and Gupta, Gourab Sen},
  title={Pause detection in continuous sign language},
  journal={International Journal of Computer Applications in Technology},
  volume={50},
  number={1/2},
  pages={75--83},
  year={2014},
  publisher={Inderscience Publishers},
  doi={10.1504/ijcat.2014.063910}
}

@inproceedings{pavlakos2019expressive,
  title={Expressive body capture: 3d hands, face, and body from a single image},
  author={Pavlakos, Georgios and Choutas, Vasileios and Ghorbani, Nima and Bolkart, Timo and Osman, Ahmed AA and Tzionas, Dimitrios and Black, Michael J},
  booktitle={Proceedings of the IEEE/CVF conference on computer vision and pattern recognition},
  pages={10975--10985},
  year={2019}
}

@article{romero2022embodied,
  title={Embodied hands: Modeling and capturing hands and bodies together},
  author={Romero, Javier and Tzionas, Dimitrios and Black, Michael J},
  journal={arXiv preprint arXiv:2201.02610},
  year={2022}
}

@article{li2017learning,
  title={Learning a model of facial shape and expression from 4D scans.},
  author={Li, Tianye and Bolkart, Timo and Black, Michael J and Li, Hao and Romero, Javier},
  journal={ACM Trans. Graph.},
  volume={36},
  number={6},
  pages={194--1},
  year={2017}
}

@inproceedings{rong2021frankmocap,
  title={Frankmocap: A monocular 3d whole-body pose estimation system via regression and integration},
  author={Rong, Yu and Shiratori, Takaaki and Joo, Hanbyul},
  booktitle={Proceedings of the IEEE/CVF International Conference on Computer Vision (ICCV) Workshops},
  pages={1749--1759},
  year={2021}
}

@inproceedings{feng2021collaborative,
  title={Collaborative regression of expressive bodies using moderation},
  author={Feng, Yao and Choutas, Vasileios and Bolkart, Timo and Tzionas, Dimitrios and Black, Michael J},
  booktitle={2021 International Conference on 3D Vision (3DV)},
  pages={792--804},
  year={2021},
  organization={IEEE}
}

@article{zhang2023pymaf,
  title={Pymaf-x: Towards well-aligned full-body model regression from monocular images},
  author={Zhang, Hongwen and Tian, Yating and Zhang, Yuxiang and Li, Mengcheng and An, Liang and Sun, Zhenan and Liu, Yebin},
  journal={IEEE Transactions on Pattern Analysis and Machine Intelligence},
  volume={45},
  number={10},
  pages={12287--12303},
  year={2023},
  publisher={IEEE}
}

@inproceedings{lin2023one,
  title={One-stage 3d whole-body mesh recovery with component aware transformer},
  author={Lin, Jing and Zeng, Ailing and Wang, Haoqian and Zhang, Lei and Li, Yu},
  booktitle={Proceedings of the IEEE/CVF Conference on Computer Vision and Pattern Recognition},
  pages={21159--21168},
  year={2023}
}

@article{cai2023smpler,
  title={Smpler-x: Scaling up expressive human pose and shape estimation},
  author={Cai, Zhongang and Yin, Wanqi and Zeng, Ailing and Wei, Chen and Sun, Qingping and Wang, Yanjun and Pang, Hui En and Mei, Haiyi and Zhang, Mingyuan and Zhang, Lei and others},
  journal={Advances in Neural Information Processing Systems},
  volume={36},
  pages={11454--11468},
  year={2023}
}

@inproceedings{sun2024aios,
  title={Aios: All-in-one-stage expressive human pose and shape estimation},
  author={Sun, Qingping and Wang, Yanjun and Zeng, Ailing and Yin, Wanqi and Wei, Chen and Wang, Wenjia and Mei, Haiyi and Leung, Chi-Sing and Liu, Ziwei and Yang, Lei and others},
  booktitle={Proceedings of the IEEE/CVF conference on computer vision and pattern recognition},
  pages={1834--1843},
  year={2024}
}

@inproceedings{zuo2025signs,
  title={Signs as tokens: A retrieval-enhanced multilingual sign language generator},
  author={Zuo, Ronglai and Potamias, Rolandos Alexandros and Ververas, Evangelos and Deng, Jiankang and Zafeiriou, Stefanos},
  booktitle={Proceedings of the IEEE/CVF International Conference on Computer Vision},
  pages={23806--23816},
  year={2025}
}

@inproceedings{kundu2026dexavatar,
  title={DexAvatar: 3D Sign Language Reconstruction with Hand and Body Pose Priors},
  author={Kundu, Kaustubh and Barua, Hrishav Bakul and Robertson-Bell, Lucy and Cai, Zhixi and Stefanov, Kalin},
  booktitle={Proceedings of the IEEE/CVF Winter Conference on Applications of Computer Vision},
  pages={5842--5852},
  year={2026}
}

@article{shen2024mm,
  title={MM-WLAuslan: multi-view multi-modal word-level Australian sign language recognition dataset},
  author={Shen, Xin and Du, Heming and Sheng, Hongwei and Wang, Shuyun and Chen, Hui and Chen, Huiqiang and Wu, Zhuojie and Du, Xiaobiao and Ying, Jiaying and Lu, Ruihan and others},
  journal={Advances in Neural Information Processing Systems},
  volume={37},
  pages={69700--69715},
  year={2024}
}

@article{van2017neural,
  title={Neural discrete representation learning},
  author={Van Den Oord, Aaron and Vinyals, Oriol and others},
  journal={Advances in neural information processing systems},
  volume={30},
  year={2017}
}

@inproceedings{hong2025salad,
  title={Salad: Skeleton-aware latent diffusion for text-driven motion generation and editing},
  author={Hong, Seokhyeon and Kim, Chaelin and Yoon, Serin and Nam, Junghyun and Cha, Sihun and Noh, Junyong},
  booktitle={Proceedings of the Computer Vision and Pattern Recognition Conference},
  pages={7158--7168},
  year={2025}
}

@inproceedings{zhang2025towards,
  title={Towards ai-driven sign language generation with non-manual markers},
  author={Zhang, Han and Shalev-Arkushin, Rotem and Baltatzis, Vasileios and Gillis, Connor and Laput, Gierad and Kushalnagar, Raja and Quandt, Lorna C and Findlater, Leah and Bedri, Abdelkareem and Lea, Colin},
  booktitle={Proceedings of the 2025 CHI Conference on Human Factors in Computing Systems},
  pages={1--26},
  year={2025}
}

@inproceedings{muller2023considerations,
  title={Considerations for meaningful sign language machine translation based on glosses},
  author={M{\"u}ller, Mathias and Jiang, Zifan and Moryossef, Amit and Rios, Annette and Ebling, Sarah},
  booktitle={Proceedings of the 61st Annual Meeting of the Association for Computational Linguistics (Volume 2: Short Papers)},
  pages={682--693},
  year={2023}
}

@inproceedings{kocabas2021pare,
  title={PARE: Part attention regressor for 3D human body estimation},
  author={Kocabas, Muhammed and Huang, Chun-Hao P and Hilliges, Otmar and Black, Michael J},
  booktitle={Proceedings of the IEEE/CVF international conference on computer vision},
  pages={11127--11137},
  year={2021}
}

@inproceedings{potamias2025wilor,
  title={Wilor: End-to-end 3d hand localization and reconstruction in-the-wild},
  author={Potamias, Rolandos Alexandros and Zhang, Jinglei and Deng, Jiankang and Zafeiriou, Stefanos},
  booktitle={Proceedings of the Computer Vision and Pattern Recognition Conference},
  pages={12242--12254},
  year={2025}
}

@inproceedings{yu2023acr,
  title={Acr: Attention collaboration-based regressor for arbitrary two-hand reconstruction},
  author={Yu, Zhengdi and Huang, Shaoli and Fang, Chen and Breckon, Toby P and Wang, Jue},
  booktitle={Proceedings of the IEEE/CVF conference on computer vision and pattern recognition},
  pages={12955--12964},
  year={2023}
}

@article{lugaresi2019mediapipe,
  title={Mediapipe: A framework for building perception pipelines},
  author={Lugaresi, Camillo and Tang, Jiuqiang and Nash, Hadon and McClanahan, Chris and Uboweja, Esha and Hays, Michael and Zhang, Fan and Chang, Chuo-Ling and Yong, Ming Guang and Lee, Juhyun and others},
  journal={arXiv preprint arXiv:1906.08172},
  year={2019}
}

@inproceedings{bragg2019sign,
  title={Sign language recognition, generation, and translation: An interdisciplinary perspective},
  author={Bragg, Danielle and Koller, Oscar and Bellard, Mary and Berke, Larwan and Boudreault, Patrick and Braffort, Annelies and Caselli, Naomi and Huenerfauth, Matt and Kacorri, Hernisa and Verhoef, Tessa and others},
  booktitle={Proceedings of the 21st international ACM SIGACCESS conference on computers and accessibility},
  pages={16--31},
  year={2019}
}

@inproceedings{athitsos2008american,
  title={The american sign language lexicon video dataset},
  author={Athitsos, Vassilis and Neidle, Carol and Sclaroff, Stan and Nash, Joan and Stefan, Alexandra and Yuan, Quan and Thangali, Ashwin},
  booktitle={2008 IEEE computer society conference on computer vision and pattern recognition workshops},
  pages={1--8},
  year={2008},
  organization={IEEE}
}

@article{joze2018ms,
  title={Ms-asl: A large-scale data set and benchmark for understanding american sign language},
  author={Joze, Hamid Reza Vaezi and Koller, Oscar},
  journal={arXiv preprint arXiv:1812.01053},
  year={2018}
}

@inproceedings{li2020word,
  title={Word-level deep sign language recognition from video: A new large-scale dataset and methods comparison},
  author={Li, Dongxu and Rodriguez, Cristian and Yu, Xin and Li, Hongdong},
  booktitle={Proceedings of the IEEE/CVF winter conference on applications of computer vision},
  pages={1459--1469},
  year={2020}
}

@inproceedings{von2008significance,
  title={The significance of facial features for automatic sign language recognition},
  author={Von Agris, Ulrich and Knorr, Moritz and Kraiss, Karl-Friedrich},
  booktitle={2008 8th IEEE international conference on automatic face \& gesture recognition},
  pages={1--6},
  year={2008},
  organization={IEEE}
}

@inproceedings{zahedi2005combination,
  title={Combination of tangent distance and an image distortion model for appearance-based sign language recognition},
  author={Zahedi, Morteza and Keysers, Daniel and Deselaers, Thomas and Ney, Hermann},
  booktitle={Joint Pattern Recognition Symposium},
  pages={401--408},
  year={2005},
  organization={Springer}
}

@inproceedings{von2010signum,
  title={Signum database: Video corpus for signer-independent continuous sign language recognition},
  author={Von Agris, Ulrich and Kraiss, Karl-Friedrich},
  booktitle={4th Workshop on the Representation and Processing of Sign Languages: Corpora and Sign Language Technologies},
  pages={243--246},
  year={2010}
}

@article{schembri2013building,
  title={Building the British sign language corpus},
  author={Schembri, Adam and Fenlon, Jordan and Rentelis, Ramas and Reynolds, Sally and Cormier, Kearsy},
  journal={Language Documentation \& Conservation},
  volume={7},
  pages={136--154},
  year={2013},
  publisher={University of Hawaii Press}
}

@inproceedings{braffort2010sign,
  title={Sign Language Corpora for Analysis, Processing and Evaluation.},
  author={Braffort, Annelies and Bolot, Laurence and Ch{\'e}telat-Pel{\'e}, Emilie and Choisier, Annick and Delorme, Maxime and Filhol, Michael and Segouat, J{\'e}r{\'e}mie and Verrecchia, Cyril and Badin, Flora and Devos, Nad{\`e}ge},
  booktitle={LREC},
  year={2010}
}

@inproceedings{efthimiou2010dicta,
  title={DICTA-SIGN: sign language recognition, generation and modelling with appllications in deaf communication},
  author={Efthimiou, E and Fotinea, S and Hanke, T and Glauert, J and Bowden, R and Braffort, Annelies and Collet, Christophe and Maragos, P and Goudenove, F},
  booktitle={International Conference on Language Resources and Evaluation},
  year={2010}
}

@inproceedings{efthimiou2012dicta,
  title={The dicta-sign wiki: Enabling web communication for the deaf},
  author={Efthimiou, Eleni and Fotinea, Stavroula-Evita and Hanke, Thomas and Glauert, John and Bowden, Richard and Braffort, Annelies and Collet, Christophe and Maragos, Petros and Lefebvre-Albaret, Fran{\c{c}}ois},
  booktitle={International conference on computers for handicapped persons},
  pages={205--212},
  year={2012},
  organization={Springer}
}

@misc{zuo2024simplebaselinespokenlanguage,
      title={A Simple Baseline for Spoken Language to Sign Language Translation with 3D Avatars}, 
      author={Ronglai Zuo and Fangyun Wei and Zenggui Chen and Brian Mak and Jiaolong Yang and Xin Tong},
      year={2024},
      eprint={2401.04730},
      archivePrefix={arXiv},
      primaryClass={cs.CV},
      url={https://arxiv.org/abs/2401.04730}, 
}

@inproceedings{fang2025signdiff,
  title={Signdiff: Diffusion model for american sign language production},
  author={Fang, Sen and Sui, Chunyu and Zhou, Yanghao and Zhang, Xuedong and Zhong, Hongbin and Tian, Yapeng and Chen, Chen},
  booktitle={2025 IEEE 19th International Conference on Automatic Face and Gesture Recognition (FG)},
  pages={1--11},
  year={2025},
  organization={IEEE}
}

@article{ho2020denoising,
  title={Denoising diffusion probabilistic models},
  author={Ho, Jonathan and Jain, Ajay and Abbeel, Pieter},
  journal={Advances in neural information processing systems},
  volume={33},
  pages={6840--6851},
  year={2020}
}

@inproceedings{rombach2022high,
  title={High-resolution image synthesis with latent diffusion models},
  author={Rombach, Robin and Blattmann, Andreas and Lorenz, Dominik and Esser, Patrick and Ommer, Bj{\"o}rn},
  booktitle={Proceedings of the IEEE/CVF conference on computer vision and pattern recognition},
  pages={10684--10695},
  year={2022}
}

@article{hochreiter1997long,
  title={Long short-term memory},
  author={Hochreiter, Sepp and Schmidhuber, J{\"u}rgen},
  journal={Neural computation},
  volume={9},
  number={8},
  pages={1735--1780},
  year={1997},
  publisher={MIT press}
}

@inproceedings{yin2024t2s,
  title={T2S-GPT: Dynamic vector quantization for autoregressive sign language production from text},
  author={Yin, Aoxiong and Li, Haoyuan and Shen, Kai and Tang, Siliang and Zhuang, Yueting},
  booktitle={Proceedings of the 62nd Annual Meeting of the Association for Computational Linguistics (Volume 1: Long Papers)},
  pages={3345--3356},
  year={2024}
}

@inproceedings{dong2024signavatar,
  title={Signavatar: Sign language 3d motion reconstruction and generation},
  author={Dong, Lu and Chaudhary, Lipisha and Xu, Fei and Wang, Xiao and Lary, Mason and Nwogu, Ifeoma},
  booktitle={2024 IEEE 18th International Conference on Automatic Face and Gesture Recognition (FG)},
  pages={1--10},
  year={2024},
  organization={IEEE}
}

@inproceedings{saunders2022signing,
  title={Signing at scale: Learning to co-articulate signs for large-scale photo-realistic sign language production},
  author={Saunders, Ben and Camgoz, Necati Cihan and Bowden, Richard},
  booktitle={Proceedings of the IEEE/CVF Conference on Computer Vision and Pattern Recognition},
  pages={5141--5151},
  year={2022}
}

@misc{jiang2025meaningful,
  title={Meaningful Pose-Based Sign Language Evaluation},
  author={Jiang, Zifan and Leong, Colin and Moryossef, Amit and G{\"o}hring, Anne and Rios, Annette and Cory, Oliver and Ivashechkin, Maksym and Tarigopula, Neha and Zhang, Biao and Sennrich, Rico and Ebling, Sarah},
  year={2025},
  eprint={2510.07453},
  archivePrefix={arXiv},
  primaryClass={cs.CL},
  url={https://arxiv.org/abs/2510.07453}
}

@inproceedings{tempfli2026how2sign,
  title={How2Sign-Synth3D: Markerless Holistic Sign Language Performance Capture and Synthetic Data for Dense Landmark Tracking},
  author={Tempfli, Levente and Huber, Stephan and Koller, Oscar and Duarte, Amanda},
  booktitle={Proceedings of the IEEE/CVF Conference on Computer Vision and Pattern Recognition (CVPR) Findings},
  pages={3501--3509},
  year={2026}
}

@inproceedings{tarres2023sign,
  title={Sign language translation from instructional videos},
  author={Tarr{\'e}s, Laia and G{\'a}llego, Gerard I and Duarte, Amanda and Torres, Jordi and Gir{\'o}-i-Nieto, Xavier},
  booktitle={Proceedings of the IEEE/CVF Conference on Computer Vision and Pattern Recognition (CVPR) Workshops},
  pages={5625--5635},
  year={2023}
}

@misc{hwang2021non,
  title={Non-Autoregressive Sign Language Production via Knowledge Distillation},
  author={Hwang, Eui Jun and Kim, Jung Ho and Cho, Suk Min and Park, Jong C.},
  year={2022},
  eprint={2208.06183},
  archivePrefix={arXiv},
  primaryClass={cs.CL},
  url={https://arxiv.org/abs/2208.06183}
}

@misc{saunders2020adversarial,
  title={Adversarial Training for Multi-Channel Sign Language Production},
  author={Saunders, Ben and Camgoz, Necati Cihan and Bowden, Richard},
  year={2020},
  eprint={2008.12405},
  archivePrefix={arXiv},
  primaryClass={cs.CV},
  url={https://arxiv.org/abs/2008.12405}
}

@inproceedings{jiang2023motiongpt,
  title={MotionGPT: Human Motion as a Foreign Language},
  author={Jiang, Biao and Chen, Xin and Liu, Wen and Yu, Jingyi and Yu, Gang and Chen, Tao},
  booktitle={Advances in Neural Information Processing Systems},
  year={2023}
}

@inproceedings{shi2022openasl,
  title={Open-Domain Sign Language Translation Learned from Online Video},
  author={Shi, Bowen and Brentari, Diane and Shakhnarovich, Greg and Livescu, Karen},
  booktitle={European Conference on Computer Vision},
  pages={582--599},
  year={2022},
  organization={Springer}
}

@misc{tang2024signiddiconicitydisentangleddiffusion,
  title={Sign-IDD: Iconicity Disentangled Diffusion for Sign Language Production},
  author={Tang, Shengeng and He, Jiayi and Guo, Dan and Wei, Yanyan and Li, Feng and Hong, Richang},
  year={2024},
  eprint={2412.13609},
  archivePrefix={arXiv},
  primaryClass={cs.CV},
  url={https://arxiv.org/abs/2412.13609}
}

@inproceedings{forte2023reconstructing,
  title={Reconstructing signing avatars from video using linguistic priors},
  author={Forte, Maria-Paola and Kulits, Peter and Huang, Chun-Hao P and Choutas, Vasileios and Tzionas, Dimitrios and Kuchenbecker, Katherine J and Black, Michael J},
  booktitle={Proceedings of the IEEE/CVF Conference on Computer Vision and Pattern Recognition},
  pages={12791--12801},
  year={2023}
}

@misc{desai2023aslcitizencommunitysourceddataset,
      title={ASL Citizen: A Community-Sourced Dataset for Advancing Isolated Sign Language Recognition}, 
      author={Aashaka Desai and Lauren Berger and Fyodor O. Minakov and Vanessa Milan and Chinmay Singh and Kriston Pumphrey and Richard E. Ladner and Hal Daumé III and Alex X. Lu and Naomi Caselli and Danielle Bragg},
      year={2023},
      eprint={2304.05934},
      archivePrefix={arXiv},
      primaryClass={cs.CV},
      url={https://arxiv.org/abs/2304.05934}, 
}

@misc{bai2018empiricalevaluationgenericconvolutional,
      title={An Empirical Evaluation of Generic Convolutional and Recurrent Networks for Sequence Modeling}, 
      author={Shaojie Bai and J. Zico Kolter and Vladlen Koltun},
      year={2018},
      eprint={1803.01271},
      archivePrefix={arXiv},
      primaryClass={cs.LG},
      url={https://arxiv.org/abs/1803.01271}, 
}

@misc{sundararajan2017axiomaticattributiondeepnetworks,
      title={Axiomatic Attribution for Deep Networks}, 
      author={Mukund Sundararajan and Ankur Taly and Qiqi Yan},
      year={2017},
      eprint={1703.01365},
      archivePrefix={arXiv},
      primaryClass={cs.LG},
      url={https://arxiv.org/abs/1703.01365}, 
}

@article{Selvaraju_2019,
   title={Grad-CAM: Visual Explanations from Deep Networks via Gradient-Based Localization},
   volume={128},
   ISSN={1573-1405},
   url={http://dx.doi.org/10.1007/s11263-019-01228-7},
   DOI={10.1007/s11263-019-01228-7},
   number={2},
   journal={International Journal of Computer Vision},
   publisher={Springer Science and Business Media LLC},
   author={Selvaraju, Ramprasaath R. and Cogswell, Michael and Das, Abhishek and Vedantam, Ramakrishna and Parikh, Devi and Batra, Dhruv},
   year={2019},
   month=Oct, pages={336–359} }

@article{morales2024ema,
  title={Exponential Moving Average of Weights in Deep Learning: Dynamics and Benefits},
  author={Morales-Brotons, Daniel and Vogels, Thijs and Hendrikx, Hadrien},
  journal={Transactions on Machine Learning Research},
  year={2024}
}

@inproceedings{wang2024uvmap,
  title={UVMap-ID: A Controllable and Personalized UV Map Generative Model},
  author={Wang, Weijie and Zhang, Jichao and Liu, Chang and Li, Xia and Xu, Xingqian and Shi, Humphrey and Sebe, Nicu and Lepri, Bruno},
  booktitle={ACM MM},
  pages={10725--10734},
  year={2024}
}

@misc{pavlakos2023reconstructinghands3dtransformers,
      title={Reconstructing Hands in 3D with Transformers}, 
      author={Georgios Pavlakos and Dandan Shan and Ilija Radosavovic and Angjoo Kanazawa and David Fouhey and Jitendra Malik},
      year={2023},
      eprint={2312.05251},
      archivePrefix={arXiv},
      primaryClass={cs.CV},
      url={https://arxiv.org/abs/2312.05251}, 
}
}

\clearpage
\appendix
\section{Cross-View Motion Curation}
\label{app:motion_curation}

This appendix records implementation details that are compressed in the main paper.
The material here explains the engineering choices behind the multi-view motion curation, tokenizer, and generation modules to ensure full reproducibility.

\subsection{Reliability-Guided Per-View Fusion}
\label{app:sub_per_view_fusion}
We define two camera views $v \in \mathcal{V}_{\mathrm{view}} = \{\mathrm{f}, \mathrm{s}\}$ for the frontal and side streams, and $h \in \{l, r\}$ denotes the hand index.
AiOS~\cite{sun2024aios} yields a stable whole-body sequence $\Theta_t^{A,v}$, whereas WiLoR~\cite{potamias2025wilor} predicts detailed hand parameters, namely $\theta^{W,v}_{t,\mathrm{w},h}$ for the wrist and $\theta^{W,v}_{t,\mathrm{fin},h}$ for the fingers (expressed in axis-angle form on $SO(3)$).

To operationalize the reliability-guided corruption mask $m^v_{t,h}$ introduced in the main paper, we formally derive its constituent evaluation metrics.
First, the per-view global wrist rotations are extracted via:

\begin{equation}
R^{A,v}_{t,\mathrm{w},h} = \mathrm{FK}_{\mathrm{w},h}(\Theta^{A,v}_t), \qquad R^{W,v}_{t,\mathrm{w},h} = \mathrm{Exp}(\theta^{W,v}_{t,\mathrm{w},h}),
\label{eq:app_wrist_rot}
\end{equation}
where $\mathrm{FK}_{\mathrm{w},h}$ represents forward kinematics applied to the AiOS tree, and $\mathrm{Exp}(\cdot)$ maps axis-angles to rotation matrices.
The hand presence validity flag is defined via an $L_1$-norm threshold as $\rho^v_{t,h} = \mathbf{1}[\|\theta^{W,v}_{t,\mathrm{w},h}\|_1 > \varepsilon]$.
The orientation discrepancy and the frame-to-frame angular velocities are quantified using the standard geodesic distance on $SO(3)$, $d_{\mathrm{geo}}(R_1, R_2) = \arccos((\mathrm{tr}(R_1^\top R_2)-1)/2)$, yielding $\omega^{W,v}_{t,h} = d_{\mathrm{geo}}(R^{W,v}_{t-1,\mathrm{w},h}, R^{W,v}_{t,\mathrm{w},h})$ and $\omega^{A,v}_{t,h} = d_{\mathrm{geo}}(R^{A,v}_{t-1,\mathrm{w},h}, R^{A,v}_{t,\mathrm{w},h})$.
These individual cues are integrated into the final per-view corruption mask:

\begin{equation}
\begin{split}
m^{v}_{t,h} = & \mathbf{1}\!\big[\neg\rho^{v}_{t,h} \vee \\
& d_{\mathrm{geo}}(R^{W,v}_{t,\mathrm{w},h},R^{A,v}_{t,\mathrm{w},h})>\tau_{\mathrm{ori}} \vee \\
& (\omega^{W,v}_{t,h}>\tau_W \wedge \omega^{A,v}_{t,h}<\tau_A)\big].
\end{split}
\label{eq:app_reliability_mask}
\end{equation}

We set $\varepsilon=10^{-6}$, $\tau_{\mathrm{ori}}=120^\circ$, $\tau_W=60^\circ$ per frame, and $\tau_A=15^\circ$ per frame; first-frame velocities are zero.
If a hand has no valid WiLoR frame, its AiOS hand parameters are retained. Clips shorter than eight frames are not processed.

To safeguard the transitions of short-lived track failures, the initial mask is temporally dilated by one frame via $\tilde m^{v}_{t,h} = \max_{|k-t|\le 1} m^{v}_{k,h}$.

For each corrupted interval $[a, b]$ flagged by $\tilde m^{v}_{t,h}=1$, let $a^- < a$ and $b^+ > b$ be the nearest valid boundary frames.
The intermediate wrist and finger joint rotations are repaired using spherical linear interpolation:
Let $R^v_{t,h,j}$ denote the pre-repair rotation (global for wrist, local for fingers) of hand joint $j$ for hand $h$ in view $v$.
\begin{equation}
\hat R^v_{t,h,j} = \mathrm{Slerp}\left(R^v_{a^-,h,j}, R^v_{b^+,h,j}; \frac{t-a^-}{b^+-a^-}\right), \quad t \in [a,b],
\label{eq:app_slerp}
\end{equation}
where $j$ indexes the structural joints within the hand kinematic chain.
If an interval touches the sequence boundaries, we fall back to nearest-valid forward or backward constant propagation.
Concurrently, when 3D hand keypoints $X^v_{t,h,j}$ are available, their coordinates are repaired via standard Euclidean interpolation: $\hat X^v_{t,h,j} = (1-\alpha_t)X^v_{a^-,h,j} + \alpha_t X^v_{b^+,h,j}$ with $\alpha_t = (t-a^-)/(b^+-a^-)$.

To re-integrate the repaired motions, the global wrist rotations $\hat R^v_{t,\mathrm{w},h}$ are mapped back into the local coordinate system relative to the wrist parent joint $\mathrm{pa}(\mathrm{w},h)$ in the stable AiOS scaffold: $\hat R^v_{t,\mathrm{w\text{-}local},h} = (R^{A,v}_{t,\mathrm{pa}(\mathrm{w},h)})^\top \hat R^v_{t,\mathrm{w},h}$.
The corresponding local axis-angle block $\hat\theta^v_{t,\mathrm{w},h}$ is recovered via the logarithmic map $\mathrm{Log}(\hat R^v_{t,\mathrm{w\text{-}local},h})$, while the repaired finger rotations are already local and directly yield $\hat\theta^v_{t,\mathrm{fin},h}$.
Finally, the complete, unified per-view SMPL-X~\cite{pavlakos2019expressive} sequence $\Theta^v_t$ is constructed by seamlessly blending the stable body trunk with the repaired hand parameters:

\begin{equation}
\Theta^{v}_t = \Theta^{A,v}_{t,\setminus\mathrm{hand}} \oplus \hat\theta^{v}_{t,\mathrm{w}} \oplus \hat\theta^{v}_{t,\mathrm{fin}},
\label{eq:app_final_fusion}
\end{equation}
where $\Theta^{A,v}_{t,\setminus\mathrm{hand}}$ denotes the baseline AiOS sequence without the hand blocks (retaining body pose, shape, and camera parameters, where \texttt{cam\_trans} stores the actual camera translation while \texttt{trans} remains a zero placeholder), and $\oplus$ indicates block-wise kinematic replacement.

\subsection{Alignment, Synchronization, Non-Sign Frames}
\label{app:sub_sync_and_clean}

Before the side view can be used as reliable verification evidence for the frontal stream, the two sequences must be spatially aligned and temporally synchronized.
To compare the streams, we use the retained fused tracking coordinates $J^v_t\in\mathbb R^{N\times3}$; these are not recomputed by forward kinematics from the final repaired pose.

Because the side-view global orientation and translation are recorded in a different camera coordinate system, we first apply a robust spatial alignment.
We compute a global root rotation correction by projecting the average framewise root discrepancy, $\Delta R_t = R^{\mathrm{f}}_{t,\mathrm{root}} (R^{\mathrm{s}}_{t,\mathrm{root}})^\top$, onto $SO(3)$, followed by a global translation shift to align the sequence centroids.
Let $\tilde J^{\mathrm{s}}_{t}$ denote the side-view joints after this coarse rigid alignment.

With spatial alignment established, we perform temporal synchronization.
As defined in the main text, we search for the optimal temporal offset $\delta^\star$ that minimizes the Mean Per Joint Position Error (MPJPE) over a predefined temporal search window $\mathcal{D}$:

\begin{equation}
\delta^\star = \arg\min_{\delta\in\mathcal D} \frac{1}{|\mathcal I_\delta| |\mathcal{B}|} \sum_{t\in\mathcal I_\delta} \sum_{j \in \mathcal{B}} \left\| J^{\mathrm{f}}_{t,j} - \tilde J^{\mathrm{s}}_{t-\delta,j} \right\|_2.
\label{eq:app_temporal_sync}
\end{equation}
Here, $\mathcal{I}_\delta$ denotes the set of valid overlapping frame indices under offset $\delta$, where a positive $\delta$ maps synchronized side-view time $t$ to source time $t-\delta$. Candidate offsets are scored strictly on $\mathcal{I}_\delta$ (with edge padding applied only when constructing the final synchronized sequence), and the joint subset $\mathcal{B}$ is restricted to the first 22 SMPL-X body joints (indices 0--21).
We intentionally exclude hands from this synchronization search because hand estimates prior to refinement are highly susceptible to blur and occlusion; including them would introduce severe noise into the temporal alignment.

Even after synchronization, continuous sign recordings contain non-lexical segments (\eg, preparation, ending motion, and pauses) that should be removed.
As introduced in the main text, we identify active signing frames using a combination of mouth activity, finger activity, and wrist elevation.

\paragraph{Non-sign activity detection and temporal editing.}
We compute three frontal activity cues over the SMPL-X joint groups $\mathcal{G}_{\mathrm{mouth}}=\{22\}$ (jaw), $\mathcal{G}_{\mathrm{wrist}}=\{20,21\}$, and $\mathcal{G}_{\mathrm{finger}}=\{25,\ldots,54\}$. For each group $g$, the raw motion heat $h^g_t$ is defined as the median frame-to-frame $\mathrm{SO}(3)$ geodesic distance across its joints (with $h^g_0=0$), smoothed by a three-frame moving average with edge padding. Let $m_g$ denote the temporal median of $h^g_t$ and $d_g=\mathrm{median}_t|h^g_t-m_g|+10^{-8}$ its median absolute deviation. We normalize each cue via
\begin{equation}
c^g_t=\operatorname{clip}\left(
\frac{h^g_t-(m_g-0.5d_g)}{2.5d_g},\, 0,\, 1\right).
\end{equation}

The wrist and finger scores undergo additional inactivity gating. For the wrists, we infer the vertical axis direction from the median wrist--shoulder displacement; if both wrists drop below their respective elbows by a margin exceeding $0.25\times$ the median shoulder width, we set $c^{\mathrm{wrist}}_t=0$. Similarly, if the raw finger heat remains below $10^{-3}$ for at least 18 consecutive frames, we set $c^{\mathrm{finger}}_t=0$. A frame is classified as active signing when all three cues exceed $0.30$:
\begin{equation}
a_t=\mathbf{1}\!\left[
c^{\mathrm{mouth}}_t\geq0.30\ \wedge\
c^{\mathrm{finger}}_t\geq0.30\ \wedge\
c^{\mathrm{wrist}}_t\geq0.30\right].
\label{eq:app_active_sign_mask}
\end{equation}

\begin{figure*}[t]
\centering
\includegraphics[width=\textwidth]{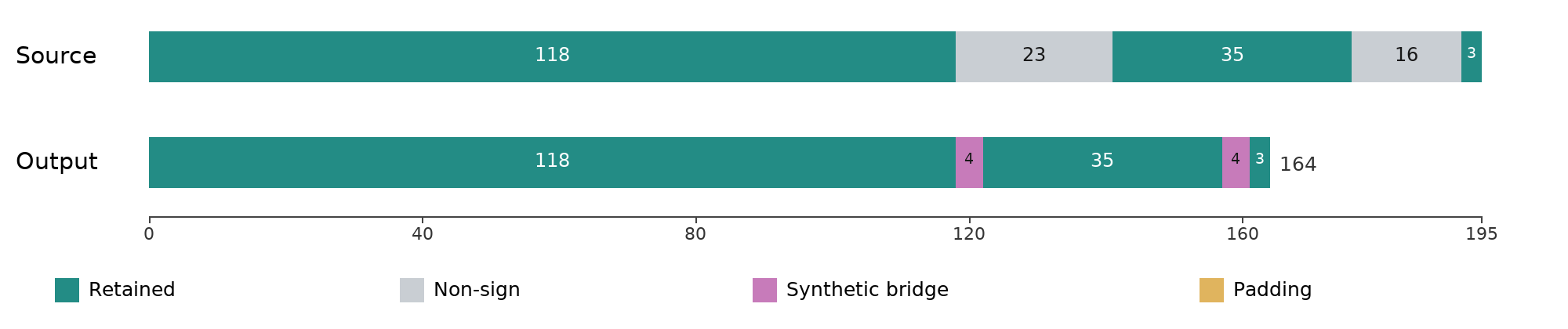}
\caption{\textbf{Timeline illustration of heuristic non-sign pruning and motion-blend transitions.} In this frontal sequence (195 source frames), removing two detected idle intervals (23 and 16 frames) leaves 156 active source frames, and inserting two 4-frame synthetic bridges yields a 164-frame continuous output sequence.}
\label{fig:afs6-idle-edit}
\end{figure*}

To preserve natural signing holds and transitions, we do not immediately truncate every inactive run ($a_t=0$). For a low-activity run starting at frame $s$, pruning begins only after a 23-frame buffer ($s+23$), and terminates up to 8 frames prior to the resumption of full signing activity ($a_t=1$). Candidate inactive intervals shorter than 12 frames ($> 8+3$) are retained, while overlapping or adjacent intervals are merged. As illustrated in Fig.~\ref{fig:afs6-idle-edit}, the detected non-signing intervals are identically excised from both views, and each cut is bridged by a 4-frame synthetic transition (using Slerp for joint rotations and linear interpolation for translational/shape parameters), with source-frame indices and synthetic-bridge masks tracked separately. Note that this kinematics-driven pruning removes prolonged idle pauses within a clip rather than recalibrating external subtitle alignments.

\subsection{Cross-View Body and Finger Repair}
\label{app:sub_cross_view_repair}

With the streams synchronized and non-signing frames removed, we cross-verify the sequences to repair abrupt tracking failures.
Hand configurations are cross-checked between the two views and their temporal neighborhoods; the final training sequence retains the frontal body reconstruction with cross-view-refined hands.

For body joints, we operate under the assumption that true, fast signing kinematics must be consistently visible across cameras.
To quantify anomalous spatial jitter, we extract the structural joint rotations $R^{v}_{t,j}$ from the fused sequence $\Theta^v_t$ and compute the frame-to-frame geodesic angular velocity $\alpha^{v}_{t,j} = d_{\mathrm{geo}}(R^{v}_{t,j}, R^{v}_{t-1,j})$.
A short temporal spike is flagged as a failure only when it manifests abruptly in the target view but is contradicted by stable, low-jitter evidence in the opposite view $\bar v$:

\begin{equation}
b^{v}_{t,j} = \mathbf{1}\!\left[ \alpha^{v}_{t,j}>\tau_{\mathrm{hi}} \wedge \alpha^{\bar v}_{t,j}\le\tau_{\mathrm{lo}} \right].
\end{equation}
When $b^{v}_{t,j}=1$, the corrupted frames in the target view are inpainted via spherical linear interpolation using the nearest valid boundary frames, identical to the boundary-aware repair in \Cref{app:sub_per_view_fusion}.
The dual-threshold design ($\tau_{\mathrm{hi}}$ and $\tau_{\mathrm{lo}}$) prevents the suppression of legitimate rapid signs.

We set $(\tau_{\mathrm{hi}},\tau_{\mathrm{lo}})=(80^\circ,30^\circ)$ for SMPL-X body joints $1,\ldots,21$ while leaving the root joint unchanged, and restrict inpainting to spike runs lasting at most two frames. Following spike repair, each rotation-vector component is denoised via a two-level \texttt{coif3} wavelet decomposition with symmetric boundary extension. We preserve the approximation coefficients and apply soft thresholding to both detail levels using the universal threshold $\hat\sigma\sqrt{2\log(\max(T,2))}$, where $\hat\sigma=\mathrm{median}(|d|)/0.6745+10^{-8}$ is estimated from the finest-scale detail coefficients $d$.

For hand articulation, the most critical residual error is the missing finger extension (\eg, a finger that should be pointing or extended collapses into a closed hand due to severe motion blur in a single view).
To execute the targeted repair mechanism introduced in the main text, we first isolate the relevant degrees of freedom.

\subsubsection{Geometric Construction of Finger Coordinate Frames}
\label{app:curation_hand_frames}

To isolate anatomical finger flexion from arbitrary local joint orientations, we construct fixed, joint-specific orthonormal coordinate frames directly from the neutral SMPL-X rest template. Let \(V_0\) denote the zero-shape template vertices and \(J_0=\mathcal{J} V_0\) the rest joints obtained via the joint regressor \(\mathcal{J}\). For each hand, we fit a reference plane via SVD to the wrist and the four non-thumb MCP joints, orienting its unit normal \(n\) downward along the body via \(-\operatorname{unit}(J_{0,12}-J_{0,0})\) to establish a consistent palmar convention.

At each finger joint \(j\), let \(y_j\) be the unit vector pointing toward its distal child joint (or the corresponding fingertip template vertex for terminal phalanges). We construct the local orthonormal frame \(T_j\) as:
\begin{equation}
\begin{split}
p_j &= \operatorname{unit}\big(n-y_j(n^\top y_j)\big),\\
x_j &= \operatorname{unit}(y_j\times p_j),\qquad
z_j = \operatorname{unit}(x_j\times y_j),\\
T_j &= [\,x_j\;y_j\;z_j\,]\in \mathrm{SO}(3),
\end{split}
\label{eq:curation_frame_construction}
\end{equation}
where the columns of \(T_j\) define the adjusted-frame axes in the SMPL-X rest coordinate system. A local joint rotation matrix \(R^{\mathrm{local}}_j\) transforms between the native and adjusted bases via similarity conjugation:
\begin{equation}
R^{\mathrm{adj}}_j=T_j^\top R^{\mathrm{local}}_jT_j,
\qquad
R^{\mathrm{local}}_j=T_jR^{\mathrm{adj}}_jT_j^\top.
\label{eq:curation_basis_change}
\end{equation}
We decompose \(R^{\mathrm{adj}}_j\) into intrinsic Euler angles as \(R^{\mathrm{adj}}_j = R_z(\gamma)R_y(\beta)R_x(\alpha)\), where the \(X\)-axis angle \(\alpha\) (denoted below as \(x_{k,i}\) for finger \(k\), joint \(i\)) captures primary finger flexion: a positive rotation bends the distal segment toward the palmar projection \(p_j\). Placing flexion on the outer \(X\)-axis avoids the middle-axis (\(\beta\)) gimbal lock singularity during standard bending motions.

\textbf{SMPL-X indexing and fingertips.} In SMPL-X, local hand pose blocks start at joint indices 25 (left) and 40 (right), ordered as index, middle, pinky, ring, and thumb (three joints per finger). Our four-finger repair operates on the non-thumb digits in anatomical order (index, middle, ring, pinky), corresponding to block offsets \((0,3,9,6)\). Each hand and joint utilizes its own dedicated frame \(T_j\). In native model finger order, the terminal fingertip vertex IDs used to define distal directions are \((4933,5058,5286,5169,5361)\) for the left hand and \((7669,7794,8022,7905,8079)\) for the right hand.

\subsubsection{Cross-View Finger Extension Repair}
\label{app:curation_finger_repair}
Single-view hand estimators frequently miss isolated finger extensions under motion blur or self-occlusion. To recover these cues across views, we evaluate the four non-thumb fingers (\(k \in \{\text{index, middle, ring, pinky}\}\)). For each finger \(k\), we extract the adjusted flexion angles \(x_{k,1}, x_{k,2}\) from its first two joints (MCP and PIP) and map them to a normalized straightness score \(S_k\in[0,1]\):
\begin{equation}
S_k=1-\operatorname{clip}\left(
\sum_{i=1}^{2}a_i\frac{|x_{k,i}|}{u_{k,i}},\, 0,\, 1\right),
\label{eq:curation_straightness}
\end{equation}
where \(a=(0.7,0.3)\) weights the proximal and intermediate joints, \(u_{k,1} \in [1.75, 1.80, 1.80, 1.70]\text{ rad}\) sets the finger-specific maximum flexion bounds for the first joint, and \(u_{k,2}=1.75\text{ rad}\) across all four fingers.

\textbf{Isolation and collapse criteria.} For a given source view, we first identify the candidate finger with the largest straightness excess over the remaining three digits:
\begin{equation*}
k^\star = \arg\max_k E_k, \qquad E_k = S_k - \mu(S_{\setminus k}).
\end{equation*}
The winning digit \(k^\star\) is confirmed as a confident isolated extension if \(S_{k^\star} > 1.15\,\mu(S_{\setminus k^\star})\) and the unbiased sample variance of the other three fingers satisfies \(\sigma^2(S_{\setminus k^\star}) < 0.2\). If confirmed, a cross-view repair is triggered in the destination view whenever the corresponding finger exhibits severe flexion collapse, defined as \(S^{\mathrm{dst}}_{k^\star} < 0.65\,S^{\mathrm{src}}_{k^\star}\).

\textbf{Bounded angular update.} Given a triggered repair, we define a target normalized flexion reduction \(d = 1.2\,E^{\mathrm{src}}_{k^\star}\) and distribute it across all three joints \(j \in \{1,2,3\}\) of digit \(k^\star\) using weights \(q=(0.7,0.7,0.5)\):
\begin{equation}
\begin{split}
\lambda &= \frac{d}{\sum_{i=1}^{2}a_iq_i/u_{k^\star,i}},\\
x^{\mathrm{dst}\prime}_{k^\star,j}
&=\operatorname{sign}(x^{\mathrm{dst}}_{k^\star,j})
\max\big(|x^{\mathrm{dst}}_{k^\star,j}|-\lambda q_j,\, 0\big).
\end{split}
\label{eq:curation_bounded_finger_update}
\end{equation}
This clamping ensures the flexion reduction never crosses zero (preventing unnatural hyperextension overshoot) while preserving the non-flexion Euler components \((\beta, \gamma)\) before mapping back to the native basis via \Cref{eq:curation_basis_change}.

\textbf{Two-pass execution.} Repair executes in two stages: (i) \emph{Cross-view pass:} we evaluate the frontal view as the source first; if it triggers a repair on a given hand and frame, the reverse (side-to-front) check is skipped for that instance; otherwise, the side view is checked to repair the frontal view. (ii) \emph{Temporal propagation pass:} each repaired event is propagated to its immediate neighbors (\(t-1\) and \(t+1\)) if the adjacent frame also fails the \(0.65\) collapse threshold and its remaining three fingers remain consistent with the source event (\(\Delta\mu < 0.15\) and \(\Delta\sigma^2 < 0.2\)).

\section{Sign-aware ST-VQVAE}
\label{app:tokenizer}

\subsection{Motion Representation and Token Layout}
The tokenizer uses the refined SMPL-X~\cite{pavlakos2019expressive} rotation representation from the curation pipeline as both input and output.
We use a selected 43-joint upper-body and hand skeleton, giving input \(X\in\mathbb R^{T\times 43\times 3}\) in axis-angle form.
After encoding and quantization, each reduced timestep produces three tokens:

\begin{equation}
z_t=\bigl(z_t^{(b)},z_t^{(l)},z_t^{(r)}\bigr),
\qquad t=1,\dots,T',
\end{equation}
corresponding to body, left, and right hand, where $T'=T/4$ after temporal downsampling.
This preserves the three-stream interface used by SOKE~\cite{zuo2025signs}-style autoregressive generation while allowing cross-part information flow before the terminal split.

\subsection{Spatio-Temporal Hierarchy}
\label{app:Skeleto-temporal Hierarchy}

Our tokenizer uses a three-level skeleton
hierarchy. The input is a selected 43 SMPL-X~\cite{pavlakos2019expressive} joint skeleton: joints \(0{:}13\) are body
joints, joints \(13{:}28\) are left-hand joints, and joints \(28{:}43\) are
right-hand joints. The pooling is not learnable. Each output node averages a
predefined group of input joints.

After the input rotations \(X\) are projected into feature space, one pooling level takes a feature tensor \(x\in\mathbb{R}^{B\times T\times J_{\mathrm{in}}\times D}\) as input. One pooling level can be written as a fixed averaging matrix
\(W\in\mathbb{R}^{J_{\mathrm{out}}\times J_{\mathrm{in}}}\):
\begin{equation}
W_{o,i} =
\begin{cases}
1/|G_o|, & i \in G_o,\\
0, & \mathrm{otherwise},
\end{cases}
\qquad
y_{b,t,o,:} = \sum_i W_{o,i} x_{b,t,i,:}.
\end{equation}
Here \(G_o\) is the joint group assigned to output node \(o\). After the first
two spatial pooling levels, we apply temporal average pooling with kernel size
2. The temporal length changes from \(T\) to \(T/2\), and then to \(T/4\). The
last level only pools space and keeps the temporal length unchanged. The final
latent skeleton has shape
\begin{equation}
U\in\mathbb{R}^{B\times T/4\times 3\times D},
\end{equation}
where the three spatial tokens are body, left hand, and right hand.

\begin{table}[t]
\centering
\scriptsize
\setlength{\tabcolsep}{2.2pt}
\caption{\textbf{Fixed skeleton hierarchy in ST-VQVAE.} Each
node is produced by average pooling over a fixed joint group. The first two levels also apply temporal average pooling with kernel size 2.}
\label{tab:h3_bodyhands_pooling}
\begin{tabular}{llll}
\toprule
\textbf{Level} & \textbf{Size/Time} & \textbf{Output Nodes} & \\
\midrule
0 & \(43\rightarrow16\) & 0: mean(0,1,2,3) & 8: mean(19,20,21) \\
  & \(T\rightarrow T/2\) & 1: mean(4,7,9,11) & 9: mean(22,23,24) \\
  & & 2: mean(5,8,10,12) & 10: mean(25,26,27) \\
  & & 3: mean(9,11,25) & 11: mean(28,29,30) \\
  & & 4: mean(10,12,40) & 12: mean(31,32,33) \\
  & & 5: mean(3,6) & 13: mean(34,35,36) \\
  & & 6: mean(13,14,15) & 14: mean(37,38,39) \\
  & & 7: mean(16,17,18) & 15: mean(40,41,42) \\
\midrule
1 & \(16\rightarrow13\) & 0: mean(prev 0,5) & 7: prev 10 \\
  & \(T/2\rightarrow T/4\) & 1: mean(prev 1,3) & 8: prev 11 \\
  & & 2: mean(prev 2,4) & 9: prev 12 \\
  & & 3: prev 6 & 10: prev 13 \\
  & & 4: prev 7 & 11: prev 14 \\
  & & 5: prev 8 & 12: prev 15 \\
  & & 6: prev 9 & \\
\midrule
2 & \(13\rightarrow3\) & body: mean(prev 0,1,2) & \\
  & \(\mathrm{unchanged}\) & left hand: mean(prev 3,4,5,6,7) & \\
  & & right hand: mean(prev 8,9,10,11,12) & \\
\bottomrule
\end{tabular}
\end{table}
\Cref{tab:h3_bodyhands_pooling} gives the exact fixed groups.

This design keeps body and hand information in one encoder before the terminal
split. At the same time, it gives separate terminal tokens for body, left hand,
and right hand, which matches the group-wise quantization used below. The
decoder uses the same fixed hierarchy in reverse: each pooled node is copied
back to the joints in its assigned group, with temporal upsampling at the two
levels that were temporally downsampled in the encoder.

We also tried other fixed hierarchies. The first type used smoother pooling,
such as \(43\rightarrow21\rightarrow9\rightarrow3\). These variants compressed
the hands earlier and gave worse results. In our setting, keeping the five
finger groups separate until the last level is more important than making the
spatial hierarchy smoother. The second type added cross-hand links, for example
letting the left-hand branch see right-hand rotations during convolution or
pooling. This also reduced reconstruction quality. We found this cross-part
information more likely to act as noise for reconstruction, although it may be
useful later in the generation model where the task is to predict coordinated
two-hand motion.

\subsection{Group-Wise Quantization}
We quantize the three terminal slots independently:

\begin{equation}
\mathcal C^{(p)}=\{e_k^{(p)}\}_{k=1}^{K_p},
\qquad
p\in\{b,l,r\},
\end{equation}
with codebook sizes:
\begin{equation}
(K_b,K_l,K_r)=(96,192,192).
\end{equation}
For each part,
\begin{equation}
z_t^{(p)}
=
\arg\min_{k\in\{1,\dots,K_p\}}
\left\|u_t^{(p)}-e_k^{(p)}\right\|_2^2,
\qquad
\tilde u_t^{(p)}=e_{z_t^{(p)}}^{(p)}.
\end{equation}
The hand codebooks are larger because sign motion contains more diverse local hand articulation than torso motion.
The codebooks are trained with EMA~\cite{morales2024ema} updates and code-usage regularization to reduce collapse.

\subsection{Training Objective}
We expand the three loss terms used in the main-text tokenizer objective. The rotation term is a masked mean-squared error over valid timesteps:

\begin{equation}
\mathcal L_{\mathrm{rot}}
=
\frac{1}{|\Omega|J}
\sum_{t\in\Omega}\sum_{j=1}^{J}
\left\|
\hat x_{t,j}-x_{t,j}
\right\|_2^2.
\end{equation}
For mesh supervision, both target and reconstruction are passed through SMPL-X:

\begin{equation}
V_t=\mathcal M(X_t),
\qquad
\hat V_t=\mathcal M(\hat X_t).
\end{equation}
We compare selected upper-body and hand vertices with a SmoothL1 penalty:

\begin{equation}
\mathcal L_{\mathrm{mesh}}
=
\frac{1}{\sum_{t\in\Omega}\sum_{q\in\mathcal V_{\mathrm{mesh}}}\omega_q}
\sum_{t\in\Omega}\sum_{q\in\mathcal V_{\mathrm{mesh}}}
\omega_q\,
\rho\!\left(\hat V_{t,q}-V_{t,q}\right).
\end{equation}
The weights \(\omega_q\) upweight hand vertices.
Without this reweighting, the mesh loss is biased toward the body and arms: they occupy a larger physical scale, contain broad surface motion, and can dominate vertex-distance averages.
This improves global/body reconstruction but can reduce the precision of finger articulation.
Since manual articulation carries much of the sign meaning, the final tokenizer deliberately trades a small amount of body/global reconstruction accuracy for stronger hand fidelity.

\subsection{Tokenizer Training Details}
We optimize the ST-VQVAE for $150$ epochs using AdamW with a learning rate of $2\times10^{-4}$ and a batch size of $32$. Once trained, the tokenizer weights are frozen, and the model is used to encode each \textbf{How2Sign-MV3D} sequence into synchronized body, left-hand, and right-hand discrete token streams for downstream generator training and retrieval.

\section{Condition and Generation}
\label{app:generation}

\subsection{Pseudo-Gloss Construction}
How2Sign~\cite{duarte2021how2sign} provides English sentences but not gold gloss annotations.
We therefore generate pseudo-glosses offline and use them as an intermediate source representation.
The pseudo-gloss is not treated as a new annotation benchmark; its purpose is to reduce the gap between natural-language word order and sign motion order before autoregressive token generation.

For few-shot prompting gloss translation, we group 20 American Sign Language English-to-gloss translation examples and 20 How2Sign English sentences into a task batch, and use GPT-5.2 API with a prompt that follows the appendix prompt style of Zhang \etal~\cite{zhang2025towards}, which reports that few-shot prompting with GPT models is effective for pseudo-gloss construction.
To better align the output glosses with our ASLCitizen~\cite{desai2023aslcitizencommunitysourceddataset} retrieval dictionary, our modification is to add the ASLCitizen vocabulary sheet as an allowed lexical reference so that generated pseudo-glosses better match the retrieval dictionary.
Although including only a single sentence per task batch reduces the dispersion of GPT-5.2's attention and improves translation quality, the token consumption for processing the dataset increases significantly. 

The resulting pseudo-gloss sequence is paired with the original English sentence during autoregressive model training.

To support reproducibility, the released code will include the prompt template, vocabulary list, and generated pseudo-glosses for the experiments.

\subsubsection{External Reference Evaluation}
\label{app:pseudogloss_validation_v3}
Table~\ref{tab:pseudogloss_validation_v2} reports the three-model
20-shot ASLG-PC12 comparison. GPT-4o is included in light of the
GPT-4o results reported by Zhang \etal~\cite{zhang2025towards}; their
dataset and evaluation settings are not identical to ours.
Among these three evaluated configurations, GPT-5.2 has the highest
reference agreement, consistent with our production choice.
This does not assert superiority over later models or supervised adaptation.
These external evaluations use one target per request and an ASLG-style
prompt, rather than replaying the vocabulary-augmented production batches.

\begin{table}[t]
\centering\small
\caption{ASLG-PC12 20-shot reference agreement on the same 200 test
sentences. The prompting approach is motivated by
Zhang \etal~\cite{zhang2025towards}; scores are our local evaluations,
not a direct reproduction of their reported scores. BLEU-4 is on a 0--1 scale.}
\label{tab:pseudogloss_validation_v2}
\begin{tabular}{lr}
\toprule
Model & BLEU-4$\uparrow$\\
\midrule
GPT-4o & .5789\\
GPT-5.2 & \textbf{.6589}\\
Qwen2.5-72B-Instruct & .5553\\
\bottomrule
\end{tabular}
\end{table}

\subsubsection{External Data and Evaluation Settings}
The local ASLG-PC12 preparation starts from 87,710 rows, removes 6,619
duplicate English/gloss pairs, and splits the remaining 81,091 pairs with
seed 42 into 64,872 training, 8,109 development, and 8,110 test records
(80/10/10). We evaluate a fixed 200-item test subset with identical
identifiers, English sentences, and references across the displayed systems.
This is a local split, not a claim to use another paper's official split.
Twenty-shot or 80-shot denotes 20 or 80 English/reference-gloss demonstration
pairs in each inference context, not 20 or 80 test targets. The later
80-shot systems use the saved 80-example training demonstration file
(selection seed 42); each request has one test target. Test references
are never included in inference prompts. The 20-shot comparisons share
the same demonstrations within that experiment.

The ASLG-style prompt asks for a single gloss line matching the reference
notation, including punctuation, \texttt{BE}, \texttt{DESC-}, and
\texttt{X-I} when appropriate. The general-gloss prompt instead follows
generic ASL-gloss conventions. This distinction is material: an
annotation-style prompt is optimized for reference agreement, not an
independent linguistic judgment. API sampling uses temperature zero;
local Qwen inference uses greedy decoding with a 128-token generation cap.
The saved API batch setup uses a 256-token output budget, while the
GPT-5.2 20-shot configuration records a 512-token budget and low reasoning.
Different prompt, output-budget, and supervision settings are explicitly
identified rather than treated as a matched-compute ranking.

BLEU-4 uses add-one smoothing on aggregate clipped 1--4-gram precisions
and the evaluator's brevity penalty, on a 0--1 scale. This is the local
implementation, not standard SacreBLEU. Higher scores indicate closer
agreement with the dataset's reference distribution; they do not directly
measure linguistic accuracy, natural signing, or downstream motion quality.

\subsubsection{Additional Few-Shot and Supervised Adaptation Study}
\label{app:pseudogloss_ft_update}

To further examine how scaling in-context examples and supervised fine-tuning (SFT) affect gloss agreement, we conduct an expanded benchmark on the 200-item ASLG-PC12 test set (Table~\ref{tab:pseudogloss_ft_update}). Because these runs use expanded supervision budgets (80-shot prompting or 5,000-sample QLoRA) and newer model checkpoints, we report them separately from the 20-shot comparison while keeping our main GPT-5.2 annotations and downstream motion experiments fixed. Table~\ref{tab:pseudogloss_ft_update} includes only configurations that returned valid outputs for all 200 test items; two thinking-enabled DeepSeek runs that timed out or returned empty strings on subsets (177/200 and 135/200 completions) were excluded to ensure a uniform evaluation denominator.

\begin{table*}[t]
\centering\small
\caption{\textbf{Expanded ASLG-PC12 comparison on the 200-item test set.} All listed configurations achieved 100\% completion across the test set. Fine-tuned (QLoRA) models leverage 5,000 in-domain training pairs, whereas API and prompting baselines operate in an 80-shot in-context regime.}
\label{tab:pseudogloss_ft_update}
\begin{tabular}{llr}
\toprule
Model & Method & BLEU-4$\uparrow$\\
\midrule
GPT-5.4 (2026-03-05) & 80-shot, API batch & .7709\\
GPT-5.5 (2026-04-23) & 80-shot, API batch & .8077\\
DeepSeek V4 Flash & 80-shot, thinking disabled & .7411\\
DeepSeek V4 Pro & 80-shot, thinking disabled & .8037\\
Qwen3.5-27B & 80-shot, general gloss prompt & .7143\\
Qwen3.5-27B & 80-shot, ASLG-style prompt & .7599\\
\midrule
Qwen2.5-7B-Instruct & QLoRA (5k SFT, 313 steps) & .9450\\
Qwen3.5-27B & QLoRA (5k SFT, 313 steps) & \textbf{.9614}\\
\bottomrule
\end{tabular}
\end{table*}

\textbf{QLoRA training setup.} Both fine-tuned models are trained on 5,000 training and 500 development pairs from ASLG-PC12 (verified to have zero exact normalized-text overlap with the 200 test items). Base models are loaded in 4-bit NF4 with double quantization and BF16 compute, applying LoRA adapters ($r=16, \alpha=32$, dropout $0.05$) to all attention (\texttt{q/k/v/o\_proj}) and MLP (\texttt{gate/up/down\_proj}) linear layers. Each sample is formatted as a zero-shot system/user prompt with the cross-entropy loss masked strictly to the assistant gloss completion (max length 512). We train for 1 epoch (313 steps) using AdamW (\texttt{adamw\_torch}, $\text{lr}=2\times 10^{-4}$, linear decay, effective batch size 16, seed 42) and evaluate the final checkpoint directly without few-shot demonstrations.

\textbf{Discussion.} On the same Qwen3.5-27B backbone, in-domain QLoRA fine-tuning achieves substantially higher ASLG-PC12 reference agreement (.9614 BLEU-4) than 80-shot prompting (.7599). While Zhang~\etal observed few-shot prompting outperforming fine-tuning under smaller data budgets on ASLLRP, our result shows that given 5,000 training pairs from a rule-based synthetic corpus like ASLG-PC12, SFT effectively internalizes corpus-specific mapping rules. However, high agreement on synthetic ASLG-PC12 glosses does not automatically translate to superior out-of-domain conditioning on continuous natural signing in How2Sign, where our frozen GPT-5.2 pipeline remains the production backbone.

\subsubsection{How2Sign Lexical Coverage}
We audit the generated How2Sign pseudo-gloss sequences against the 2,300-sign ASLCitizen vocabulary (Table~\ref{tab:pseudogloss_coverage_v2}). \emph{Exact coverage} measures strict token-level matches, whereas \emph{Loose coverage} strips peripheral punctuation and hyphens after uppercasing. \emph{Rows with match} denotes the proportion of sentences containing at least one matched dictionary sign.

\begin{table*}[t]
\centering\scriptsize
\caption{\textbf{How2Sign pseudo-gloss formatting and ASLCitizen lexical coverage audit.}}
\label{tab:pseudogloss_coverage_v2}
\begin{tabular}{lrrrrrrr}
\toprule
Split & Rows & Empty & Multiline & Avg. tokens & Exact coverage & Loose coverage & Rows with match\\
\midrule
Train & 31,165 & 0 & 0 & 11.90 & 59.22\% & 59.84\% & 98.10\%\\
Val & 1,741 & 0 & 0 & 12.00 & 57.77\% & 58.39\% & 97.30\%\\
Test & 2,357 & 0 & 0 & 12.17 & 56.09\% & 56.68\% & 97.62\%\\
All & 35,263 & 0 & 0 & 11.92 & 58.94\% & 59.55\% & 98.03\%\\
\bottomrule
\end{tabular}
\end{table*}

All 35,263 sentences yield well-formed, single-line outputs without empty strings. Across the full dataset, 59.55\% of generated tokens match the dictionary under loose matching, and 98.03\% of sentences trigger at least one retrieved sign prompt. Unmatched tokens primarily stem from open-vocabulary entities, including fingerspelling markers or single-letter tokens (present in 2,817 sentences), numeric digits (1,741 sentences), proper nouns, technical terms, and compound variants (plus 33 rows containing residual lowercase characters).

\subsubsection{Reference Disagreements and Evaluation Nuances}
\begin{table*}[t]
\centering\scriptsize
\caption{\textbf{Qualitative comparison of English inputs, NCSLGR reference glosses, and GPT-5.2 predictions,} illustrating differences in compression, lexical choice, and grammatical marking conventions.}
\label{tab:pseudogloss_errors_v2}
\begin{tabular}{p{.22\textwidth}p{.20\textwidth}p{.29\textwidth}p{.19\textwidth}}
\toprule
English & Dataset reference & Prediction & Structural observation\\
\midrule
I would drive across. & \texttt{DRIVE} &
\texttt{IX-1P DRIVE CROSS} & Reference is highly compact; prediction explicitly encodes subject pronoun and path.\\
Does the teacher like chocolate? & \texttt{TEACH+AGENT LIKE CHOCOLATE} &
\texttt{CHOCOLATE REALLY LEARN+AGENT LIKE Q} &
Prediction applies topic-fronting and an explicit question marker (\texttt{Q}), but substitutes \texttt{LEARN+AGENT}.\\
What did John buy yesterday? & \texttt{JOHN BUY YESTERDAY} &
\texttt{YESTERDAY JOHN BUY WHAT PART HOLD} &
Prediction fronts temporal adverb and appends explicit WH-question and aspectual markers.\\
\bottomrule
\end{tabular}
\end{table*}

Standard n-gram metrics such as BLEU-4 quantify surface alignment against a specific corpus convention rather than absolute linguistic validity. As illustrated by the qualitative NCSLGR comparisons in Table~\ref{tab:pseudogloss_errors_v2}, reference glosses are sometimes highly compressed (\eg, omitting subject pronouns or path modifiers that the LLM explicitly predicts), whereas LLM predictions can occasionally introduce unsupported grammatical markers (\eg, \texttt{LEARN+AGENT} instead of \texttt{TEACH+AGENT}, or trailing discourse particles). Because natural sign language allows multiple valid realizations depending on visual context and annotation guidelines, text-only reference comparisons have inherent limits. Ultimately, because How2Sign lacks ground-truth gloss annotations, we treat LLM-generated pseudo-glosses as structured semantic intermediates for motion conditioning rather than formal linguistic transcriptions.

\subsection{Gloss Adapter}
The gloss adapter maps the pseudo-gloss token sequence into the hidden space of the generator.
Given gloss tokens \(g=(g_1,\ldots,g_M)\), we compute:

\begin{equation}
\begin{aligned}
E_g&=\mathrm{Emb}_g(g),\\
H_g&=\mathrm{Adapter}(E_g+\mathrm{PE}_g),\\
\hat H_g&=W_gH_g.
\end{aligned}
\end{equation}
The adapter consists of learned gloss embeddings, learned positional embeddings, a lightweight Transformer encoder, and a linear projection.
This allows the generator to condition on English semantics and pseudo-gloss structure without replacing the existing SOKE-style backbone.

\subsection{Retrieval Dictionary}
\label{app:retrieval_dictionary}
The retrieval dictionary is built from isolated sign motion.
Each isolated sign sequence is converted into the same SMPL-X representation and encoded by the frozen ST-VQVAE tokenizer:

\begin{equation}
Z_w \in \mathbb{N}^{L_w\times 3},
\end{equation}
where $L_w$ is the token length of isolated sign entry $w$, and the three columns are body, left-, and right-hand tokens.
Each entry stores the gloss label, token matrix, and metadata used for informative-window selection.
Gloss matching is performed after normalization and alias handling.
If no pseudo-gloss token matches the dictionary, the model falls back to gloss-only conditioning.

\subsection{Auxiliary Isolated Sign Classifier}
\label{sec:app_aux_slr}

We train an auxiliary TCN-based isolated sign classifier on ASLCitizen~\cite{desai2023aslcitizencommunitysourceddataset} to extract temporal saliency from isolated motion sequences. This classifier is not intended as a sign language recognition contribution; it is only used to estimate which frames are informative within each isolated sign. All variable-length sequences are temporally resized to \(t_{\mathrm{fix}}=96\). We use position and velocity features from 56 selected \texttt{smpl\_kp3d} joints and 43 selected \texttt{poses} joints, giving
\begin{equation}
d_{\mathrm{in}}=(56+43)\times 6=594 .
\end{equation}
The classifier is a lightweight TCN with three temporal convolution layers:
\[
594 \rightarrow 256 \rightarrow 256 \rightarrow 256,
\]
using kernel sizes \(5,5,3\), BatchNorm, ReLU, and dropout 0.5 after the first two layers. Temporal average pooling is followed by a linear classifier over 2730 training classes.

We train three out-of-fold classifiers on folds BC, AC, and AB, and validate them on folds A, B, and C, respectively. Each sample is explained by the classifier that was not trained on that sample. The training uses batch size 64, 300 epochs, learning rate \(5\times10^{-4}\), weight decay \(5\times10^{-4}\), label smoothing 0.1, early stopping patience 5 with minimum delta \(10^{-4}\), and temporal scale augmentation in \([0.5,1.0]\) with edge padding. The trained classifiers reach above 95\% top-5 accuracy, which is sufficient for our saliency extraction.

For class score \(S_c\), input \(x\), and baseline \(x'\), Integrated Gradients computes
\begin{equation}
A
=
(x-x')\odot
\int_{\alpha=0}^{1}
\frac{\partial S_c(x'+\alpha(x-x'))}{\partial x}
\,d\alpha .
\end{equation}
We reduce the attribution tensor over joints and feature channels to obtain a temporal importance curve:
\begin{equation}
a_t=\mathrm{Reduce}_{j,d}\left(A_{t,j,d}\right).
\end{equation}
The retrieval module uses the normalized temporal saliency curve after joint averaging; the full joint-time map is saved only for analysis.

\subsection{Keyframe-Aware Retrieval Window}
\label{app:token_aligned_window}

The saliency curve is computed on video frames, but retrieval prompts are used
after VQ tokenization. Since the tokenizer has a temporal compression factor of
4, we first average frame-level saliency within each 4-frame token bin. This
gives a token-level saliency curve.

We then choose a short retrieval window from this curve. We use \(\tau=0.20\) to identify candidate high-saliency tokens, and then select the fixed 10-token window from the highest-weighted region of the token-level saliency curve. If no token passes the threshold, we center the window on the highest-scoring token instead. The main experiments use a fixed 10-token window. This keeps the
retrieval prompt length constant while moving the prompt to the most informative
part of the sign.

\paragraph{Clip-specific selection and sign boundaries.}
The retrieval window is selected from each isolated clip's own attribution curve rather than a shared Gaussian envelope, allowing early or late salient regions to anchor the window. Note that this dictionary-side prompt selection is distinct from continuous non-sign removal, and functions as a compact motion cue rather than a strict linguistic boundary detector for multi-phase signs.

\begin{table*}[t]
\centering
\scriptsize
\setlength{\tabcolsep}{2.4pt}
\caption{\textbf{Retrieval prompt diagnostic on How2Sign-MV3D.} E0 is the
setting used in the main experiments. Full-sequence weighting does not improve
hand reconstruction over the fixed 10-token crop.}
\label{tab:app_rag_weight}
\begin{tabular*}{\textwidth}{@{\extracolsep{\fill}}llcccccc@{}}
\toprule
Setting & Retrieval prompt
& \multicolumn{3}{c}{DTW-JPE$\downarrow$}
& \multicolumn{3}{c}{DTW-PA-JPE$\downarrow$} \\
\cmidrule(lr){3-5}\cmidrule(l){6-8}
& & Body & LHand & RHand & Body & LHand & RHand \\
\midrule
E0 fixed crop, \(\tau=.20\) & 10-token crop
& 4.983 & 8.197 & \textbf{7.697} & 4.577 & \textbf{1.814} & \textbf{1.906} \\
E1 fixed crop, \(\tau=.14\) & 10-token crop
& 5.137 & \textbf{8.192} & 8.179 & \textbf{4.570} & 1.983 & 1.996 \\
E2 saliency scale & full sequence
& 5.159 & 8.356 & 8.434 & 4.685 & 1.983 & 1.998 \\
E3 weight embedding & full sequence
& 5.261 & 8.343 & 8.390 & 4.744 & 1.987 & 1.999 \\
E4 scale + embedding & full sequence
& 5.189 & 8.207 & 8.209 & 4.659 & 1.944 & 1.979 \\
E5 scale + embedding + local position & full sequence
& 5.132 & 8.216 & 8.234 & 4.632 & 1.965 & 1.969 \\
\bottomrule
\end{tabular*}
\end{table*}    
We also compare this fixed-window design with softer alternatives in
\Cref{tab:app_rag_weight}. E1 keeps the same crop length but lowers the trigger
threshold, while E2--E5 retain the full isolated-sign sequence and inject
saliency through scaling, learned weight embeddings, or local position
embeddings. These alternatives increase the amount of retrieval context, but the
results favor a short explicit lexical window for the hand-aligned metrics most
affected by retrieval prompts.

\subsection{Generator Training Details}
\label{app:generation_training_details}

We train the generator on How2Sign-MV3D motion tokens. The model follows the
SOKE~\cite{zuo2025signs} framework and language backbone. We fully finetune it during
training. The motion tokenizer is frozen.
The input condition uses the pseudo-gloss sequence and ASLCitizen retrieval dictionary described in \Cref{app:retrieval_dictionary}.

The generator is trained with teacher-forced cross-entropy over the three token
heads. At inference time, it autoregressively predicts body, left-hand, and
right-hand tokens, then decodes them through the frozen ST-VQVAE decoder. Unless
otherwise specified, decoding is greedy with a maximum length of \(100\) tokens.

Using the \textbf{How2Sign-MV3D} train/validation/test split described in Section~\ref{sec:experimental_setup}, we train the generator for $150$ epochs with a batch size of $48$ using AdamW ($\beta_1=0.9, \beta_2=0.99$, zero weight decay) and a cosine learning-rate schedule decaying from $2\times 10^{-4}$ to a minimum of $10^{-6}$.

\section{Evaluation Details}
\label{app:training}

\subsection{SGNify-MV Reconstruction Diagnostic}
\label{app:sgnify_mv}

We build \textbf{SGNify-MV} to test our cross-view curation pipeline under known 3D geometry.
SGNify provides monocular videos and ground-truth meshes, but no side-view video.
We are not aware of a public sign language benchmark that provides both multi-view video and ground-truth SMPL-X meshes.
We therefore render frontal and side-view videos from the SGNify~\cite{forte2023reconstructing} ground-truth meshes in Blender.

The original SGNify~\cite{forte2023reconstructing} videos are clean $30$-FPS isolated signs with slow and standard motion, yet How2Sign~\cite{duarte2021how2sign} videos are $24$-FPS continuous sign language.
To make the rendered benchmark closer to the failures seen in How2Sign, we render SGNify~\cite{forte2023reconstructing} 3D annotation at $24$ FPS, add motion blur, and apply random temporal offsets between views.
SGNify stores each frame as a separate OBJ mesh, but the topology and vertex ordering are fixed.
We use this to build Blender Shape Keys, convert the discrete mesh sequence into continuous deformation, and sample different views at slightly different times.
UV and Texture made by UVMap-ID~\cite{wang2024uvmap}
\begin{figure*}[t]
\centering
\includegraphics[width=\textwidth]{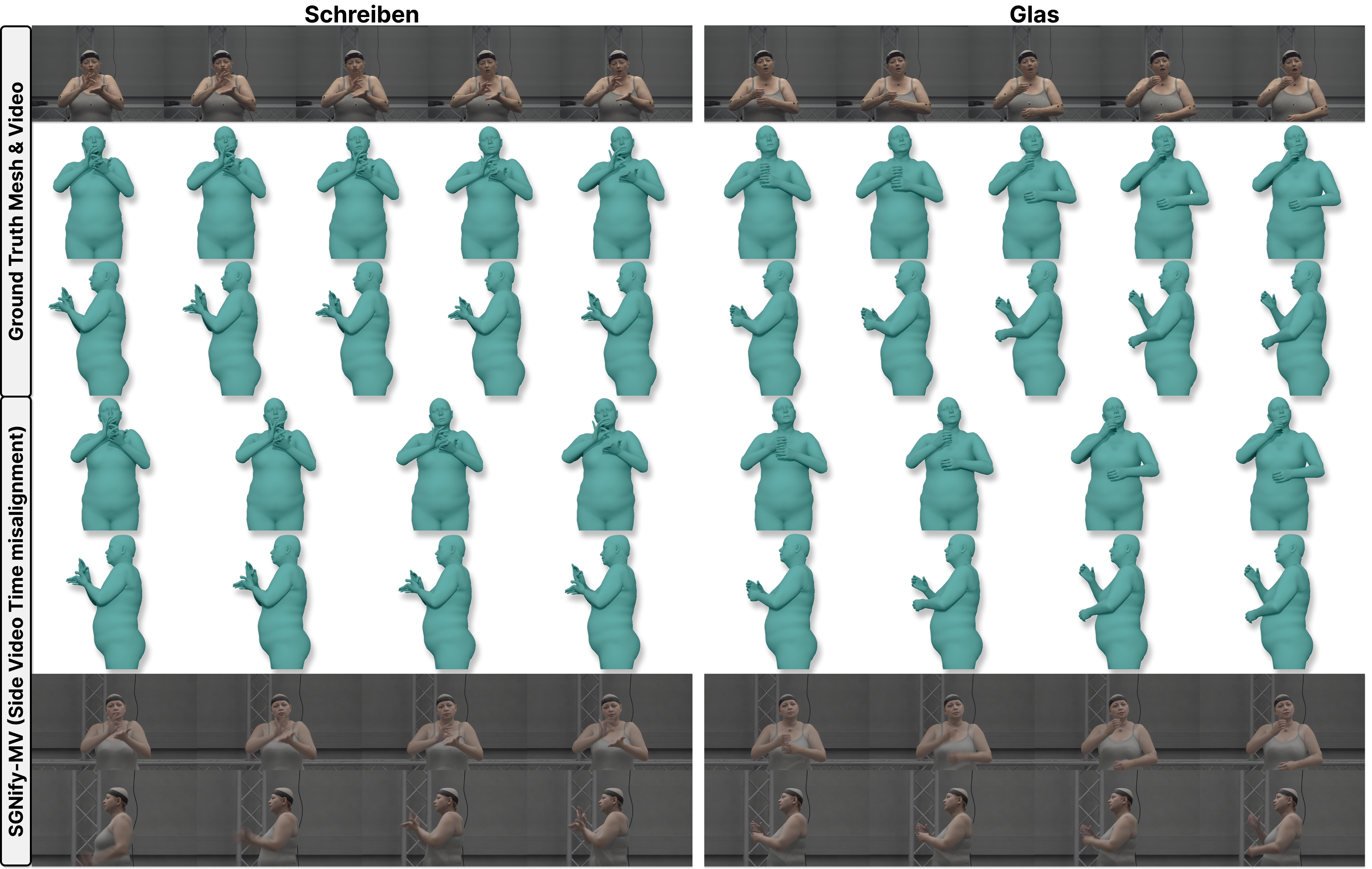}
\caption{\textbf{SGNify-MV visualization.} Examples for two SGNify signs. We
show the original SGNify~\cite{forte2023reconstructing} video, ground-truth mesh sequence, and our rendered low FPS frontal/side videos with controlled temporal mismatch and motion blur.}
\label{fig:sgnify_mv_visual}
\end{figure*}

\begin{table}[!t]
\centering
\caption{\textbf{TR-V2V results on SGNify and SGNify-MV.} We report TR-V2V error (mm) for UBody-F and both hands. The first block follows the DexAvatar~\cite{kundu2026dexavatar} SGNify evaluation protocol. The remaining blocks report our results using DexAvatar's evaluation code.}
\label{tab:sgnify_mv}
\scriptsize
\setlength{\tabcolsep}{2.4pt}
\renewcommand{\arraystretch}{0.95}
\begin{tabular*}{\columnwidth}{@{\extracolsep{\fill}}llccc@{}}
\toprule
\textbf{Setting/Method} & \textbf{Input} & \textbf{UBody-F $\downarrow$} & \textbf{LHand $\downarrow$} & \textbf{RHand $\downarrow$} \\
\midrule
\multicolumn{5}{l}{\emph{Real SGNify, DexAvatar protocol}} \\\\
OSX~\cite{lin2023one} & Front & 47.32 & 18.34 & 18.12 \\
Neural Sign Actors~\cite{baltatzis2024neural} & Front & 46.42 & 16.17 & 15.23 \\
EVA*~\cite{hu2024expressivegaussian} & Front & 40.38 & 13.73 & 13.68 \\
DexAvatar~\cite{kundu2026dexavatar} & Front & \textbf{30.13} & \textbf{13.53} & \textbf{13.08} \\
\midrule
\multicolumn{5}{l}{\emph{Real SGNify}} \\\\
OSX~\cite{lin2023one} & Front & 45.38 & 34.39 & 30.54 \\
AiOS~\cite{sun2024aios} & Front & 37.42 & 35.67 & 36.88 \\
AiOS+WiLoR~\cite{sun2024aios,potamias2025wilor} & Front & \textbf{34.61} & 13.11 & 12.45 \\
Ours one-view & Front & 34.62 & \textbf{12.87} & \textbf{12.34} \\
\midrule
\multicolumn{5}{l}{\emph{Rendered SGNify-MV}} \\\\
OSX~\cite{lin2023one} & Front & 40.50 & 32.48 & 29.92 \\
AiOS~\cite{sun2024aios} & Front & 38.36 & 38.88 & 42.80 \\
AiOS+WiLoR~\cite{sun2024aios,potamias2025wilor} & Front & 33.50 & 17.36 & 21.38 \\
Ours one-view & Front & \textbf{32.77} & \textbf{14.94} & \textbf{17.86} \\
\specialrule{0.08em}{0.6ex}{0.6ex}
Ours cross-view & Front+Side & 32.80 & 14.95 & 17.88 \\
\bottomrule
\end{tabular*}
\end{table}

TR-V2V shown in \Cref{tab:sgnify_mv} is our main reconstruction diagnostic because it keeps the hand in the body coordinate space.
This matters for signing: the same finger shape can have a different meaning when it appears near a different body part.
Part-aligned hand errors are useful for local handshape analysis, but they remove this global placement information.

On real-front SGNify~\cite{forte2023reconstructing}, one-view repair slightly improves the hand scores.
On rendered SGNify-MV, cross-view repair remains close to one-view repair rather
than improving it, suggesting that the side-view evidence is limited by temporal
mismatch in this controlled rendered setting.

\begin{table}[t]
\centering
\footnotesize
\caption{\textbf{Hand motion dynamics.} We compare wrist and finger motion measured from GT SMPL-X meshes for SGNify and pseudo SMPL-X meshes for the other datasets. F/W Ratio is the average global finger motion divided by the average global wrist motion. Local Finger subtracts the wrist position before measuring finger movement. All motion values are speeds in the SMPL-X coordinate scale, computed as frame-to-frame 3D joint displacement multiplied by frame rate.}
\label{tab:hand_motion_dynamics}

\begingroup
\setlength{\tabcolsep}{2pt}
\renewcommand{\arraystretch}{1.1}

\begin{tabular*}{\columnwidth}{@{\extracolsep{\fill}}lcccc@{}}
\toprule
\textbf{Dataset} &
\textbf{Wrist} &
\textbf{Finger} &
\textbf{F/W Ratio} &
\shortstack{\textbf{Local}\\\textbf{Finger}} \\
\midrule
ASLCitizen & 0.301 & 0.408 & 1.36 & 0.188 \\
How2Sign & 0.252 & 0.471 & 1.87 & 0.308 \\
MM-WLAuslan & 0.382 & 0.547 & 1.43 & 0.220 \\
SGNify & 0.466 & 0.535 & 1.15 & 0.124 \\
SGNify(central) & 0.383 & 0.510 & 1.33 & 0.220 \\
\bottomrule
\end{tabular*}

\endgroup
\end{table}

\Cref{tab:hand_motion_dynamics} helps explain why this diagnostic result should
not be read as the full behavior on continuous How2Sign~\cite{duarte2021how2sign}.
SGNify has weaker local finger dynamics than How2Sign: the finger-to-wrist ratio
is 1.15 for SGNify-MV and 1.87 for How2Sign, and local finger motion is 0.124
versus 0.308. Even after trimming SGNify to its central signing interval, its
local finger motion remains below How2Sign. SGNify-MV therefore provides
controlled geometry and useful failure cases, but it does not fully match the
faster, more finger-active continuous signing regime where How2Sign-MV3D is used
for model training.

\begin{table}[!t]
\centering
\caption{\textbf{MPVPE-PA results on SGNify and SGNify-MV.} We report MPVPE-PA (mm) for UBody-F and both hands. The first block follows the How2Sign-Synth3D~\cite{tempfli2026how2sign} SGNify evaluation protocol. The remaining blocks report our results using SOKE~\cite{zuo2025signs}'s evaluation code.}
\label{tab:sgnify_mv_mpvpe_pa}
\scriptsize
\setlength{\tabcolsep}{2.4pt}
\renewcommand{\arraystretch}{0.95}
\begin{tabular*}{\columnwidth}{@{\extracolsep{\fill}}llccc@{}}
\toprule
\textbf{Setting/Method} & \textbf{Input} & \textbf{UBody-F $\downarrow$} & \textbf{LHand $\downarrow$} & \textbf{RHand $\downarrow$} \\
\midrule
\multicolumn{5}{l}{\emph{Real SGNify front, H2S-Synth3D protocol}} \\\\
OSX~\cite{lin2023one} & Front & 47.32 & 18.34 & 18.12 \\
SGNify~\cite{forte2023reconstructing} & Front & 55.63 & 19.22 & 17.50 \\
Neural Sign Actors~\cite{baltatzis2024neural} & Front & 46.42 & 16.17 & 15.23 \\
SignAvatars~\cite{yu2024signavatars} & Front & 38.50 & 15.56 & 13.28 \\
SOKE~\cite{zuo2025signs} & Front & 46.73 & \textbf{10.55} & 8.94 \\
H2S-Synth3D~\cite{tempfli2026how2sign} & Front & \textbf{20.37} & 10.63 & \textbf{8.83} \\
\midrule
\multicolumn{5}{l}{\emph{Real SGNify}} \\\\
Ours one-view & Front & 22.67 & \textbf{8.47} & \textbf{9.06} \\
\midrule
\multicolumn{5}{l}{\emph{Rendered SGNify-MV}} \\\\
Ours one-view & Front & 17.33 & 9.46 & 11.83 \\
\specialrule{0.08em}{0.6ex}{0.6ex}
Ours cross-view & Front+Side & \textbf{17.32} & 9.55 & 11.73 \\
\bottomrule
\end{tabular*}
\end{table}

Although part-aligned MPVPE-PA is not sensitive to hand direction and position errors, we also report MPVPE-PA in \Cref{tab:sgnify_mv_mpvpe_pa} because recent SGNify tables, including How2Sign-Synth3D~\cite{tempfli2026how2sign}, report part-aligned MPVPE-PA. We also avoid merging published rows into our table: some SGNify values appear under different metric names across papers, so a single mixed leaderboard would be ambiguous.

\subsection{Back-Translation BLEU Diagnostic}
\label{app:bt_bleu}

We do not report back-translation BLEU as a main metric. Recent pose-based SLP evaluation work~\cite{jiang2025meaningful,tarres2023sign,shi2022openasl} observes that back translation style metrics require a pose-to-text translation model for the relevant language direction and dataset, but the evaluator code, weights, and upper bounds are rarely standardized or released~\cite{baltatzis2024neural,zuo2025signs}; training such a model for each dataset is itself a costly extra task, and human evaluation is still needed for semantic validity~\cite{jiang2025meaningful}.
This is especially problematic on open-domain ASL data: How2Sign translation remains difficult even for dedicated sign language translation systems~\cite{tarres2023sign}, and large online-video resources such as OpenASL~\cite{shi2022openasl} still show the difficulty of robust ASL-to-English translation in unconstrained settings.

We also ran internal diagnostics with open back translation style evaluators. A SIGN-IDD-style open SLT model~\cite{tang2024signiddiconicitydisentangleddiffusion}
trained on How2Sign reached less than 2 BLEU-4 in our setup. Replacing its pose input module with our ST-VQVAE representation, connecting it through a Transformer adaptor, and decoding with T5 text models improved the best BLEU-4 to roughly 6, but the model still showed partial collapse. These results are useful diagnostics, but they are far below the reliability needed to rank SLP systems by private back translation BLEU alone.

\begin{figure*}[t]
\centering
\includegraphics[width=\textwidth]{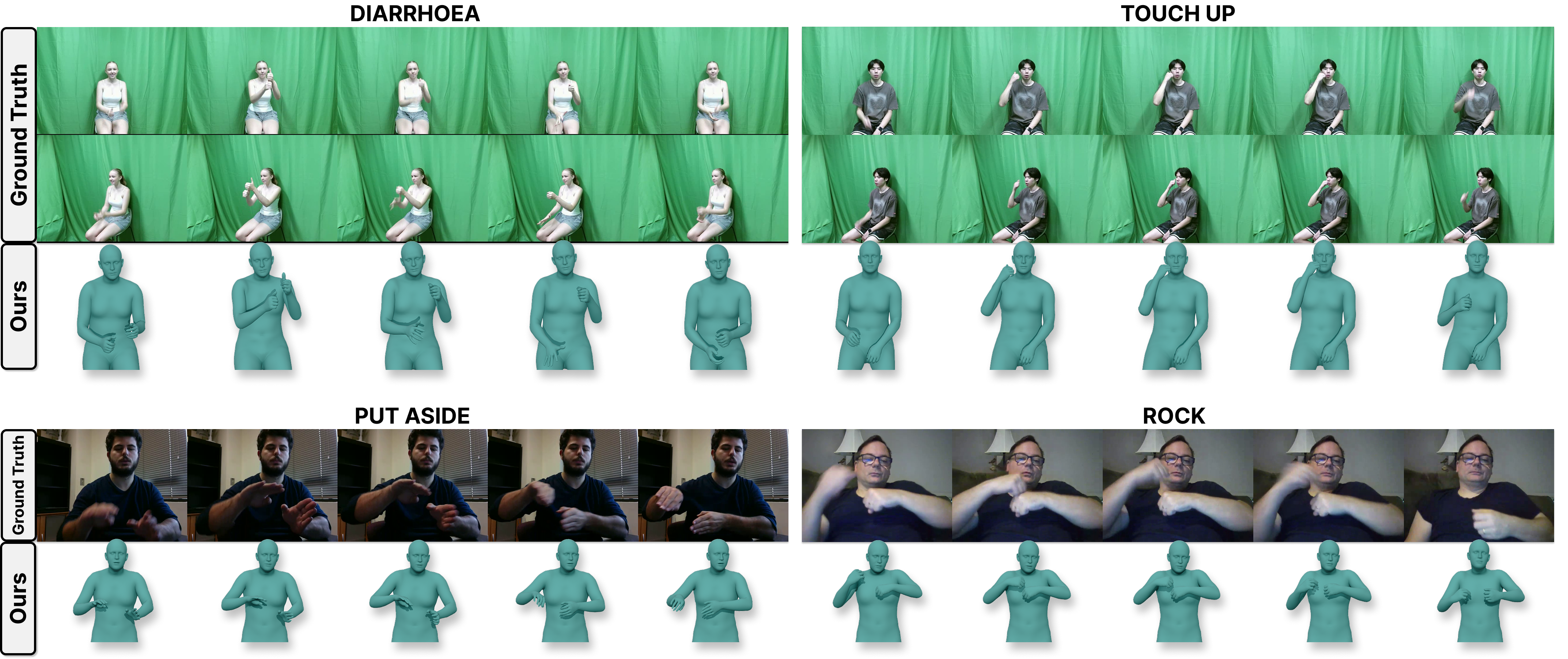}
\caption{\textbf{MM-WLAuslan-MV3D and ASLCitizen-3D visualization.} Examples for two signs for each video dataset. We
show the original ASLCitizen~\cite{desai2023aslcitizencommunitysourceddataset} and MM-WLAuslan~\cite{shen2024mm} videos, with our 3D annotations below.}
\label{fig:MMW_ASL_visual}
\end{figure*}
\subsection{ASLCitizen-3D and MM-WLAuslan-MV3D visualization}
We processed ASLCitizen~\cite{desai2023aslcitizencommunitysourceddataset} and MM-WLAuslan~\cite{shen2024mm} using our single-view/cross-view curation pipeline to visually evaluate our pipeline, and also provide additional 3D sign language annotations for future work.

\section{Front--Side Detection Availability and Reliability}
\label{app:afs6-detection}

We audit the raw AiOS and WiLoR outputs across the 31,474 jointly available paired clips (from the 35,241-clip manifest) prior to non-sign removal, interpolation, cross-view repair, and synthetic bridging. All statistics are frame-weighted, and left/right labels denote estimator-assigned anatomical hands rather than image-space sides. Because How2Sign lacks 3D ground truth and manual visibility annotations, these metrics characterize raw estimator behavior and heuristic filtering rates rather than ground-truth detection recall.

\begin{table*}[t]
\centering
\small
\caption{\textbf{Raw detection availability and heuristic rejection rates.} \emph{No box} and \emph{Rule rejected} are normalized by all decoded frames, whereas \emph{Rejected / finite} uses valid finite fits as the denominator.}
\label{tab:afs6-hand-availability}
\begin{tabular}{llllll}
\hline
View & Hand & Frames & No box (\%) & Rule rejected (\%) & Rejected / finite (\%) \\
\hline
front & left & 5,508,797 & 2.28 & 4.15 & 1.92 \\
front & right & 5,508,797 & 0.67 & 2.49 & 1.83 \\
side & left & 5,508,715 & 8.41 & 15.39 & 7.62 \\
side & right & 5,508,715 & 23.06 & 34.32 & 14.64 \\
\hline
\end{tabular}
\end{table*}

In Table~\ref{tab:afs6-hand-availability}, \emph{No box} indicates frames where no hand bounding box meets the 0.3 confidence threshold (encompassing both missed detections and genuine occlusions or out-of-frame movements). For all detected boxes in this cohort, the exported pose and keypoint arrays are finite. 

Beyond box detection, our native merge rule rejects a hand fit if: (i) its wrist axis-angle $L_1$ norm is $\le 10^{-6}$ (a near-zero-rotation heuristic distinct from the detector box mask), (ii) the AiOS--WiLoR wrist orientation disagreement exceeds $120^\circ$, or (iii) WiLoR changes by $>60^\circ$/frame while AiOS changes by $<15^\circ$/frame. Because these failure modes overlap, we report their union after a one-frame temporal dilation. Among finite raw fits, this rule rejects an additional 1.92\%, 1.83\%, 7.62\%, and 14.64\% of frames for front-left, front-right, side-left, and side-right hands, respectively. Note that an inaccurate AiOS wrist reference can occasionally trigger a false orientation rejection on a plausible WiLoR hand fit.

\begin{figure*}[!t]
\centering
\includegraphics[width=\textwidth]{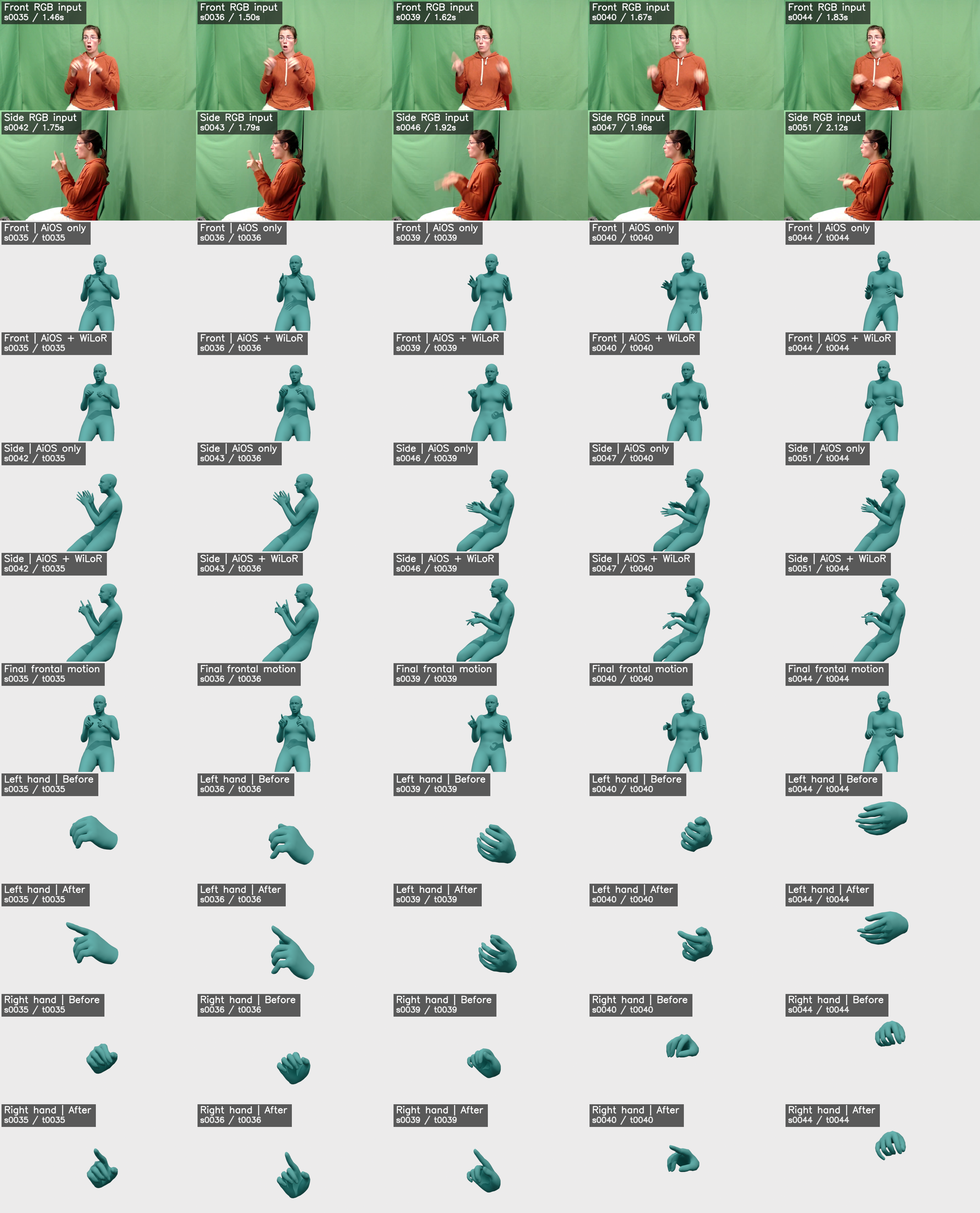}
\caption{\textbf{Stage-by-stage pipeline visualization on clip \texttt{1xl3MOjKX84\_2-8-rgb}.} From top to bottom: synchronized frontal and side RGB frames, native-view AiOS-only meshes, and per-view AiOS+WiLoR merged meshes (incorporating single-view reliability filtering and interpolation prior to cross-view synchronization and finger repair). Note the visible torso tilt in the side-view body meshes, which motivates retaining the frontal body backbone. The bottom rows highlight the frontal output sequence with cross-view-refined hands, illustrating local finger configurations before and after cross-view repair across five updated timestamps (with root, body, and shape held fixed for visual comparison).}
\label{fig:afs6-author-finger-repair}
\end{figure*}

For the body, AiOS outputs dense per-frame meshes; however, side-view body fits frequently exhibit under-fitting and implausible torso tilt (see Fig.~\ref{fig:afs6-author-finger-repair}). Consequently, we retain the frontal body reconstruction as the kinematic backbone while performing bidirectional cross-view refinement for the hands. Specifically, frontal and side hand configurations are mutually validated alongside their temporal neighbors, allowing reliable side-view observations to correct occluded frontal hands and vice versa. Finally, in our raw-body temporal alignment diagnostic, the offset search reaches the $\pm 8$-frame boundary in 40.40\% of clip pairs.

\begin{figure*}[t]
\centering
\includegraphics[width=\textwidth]{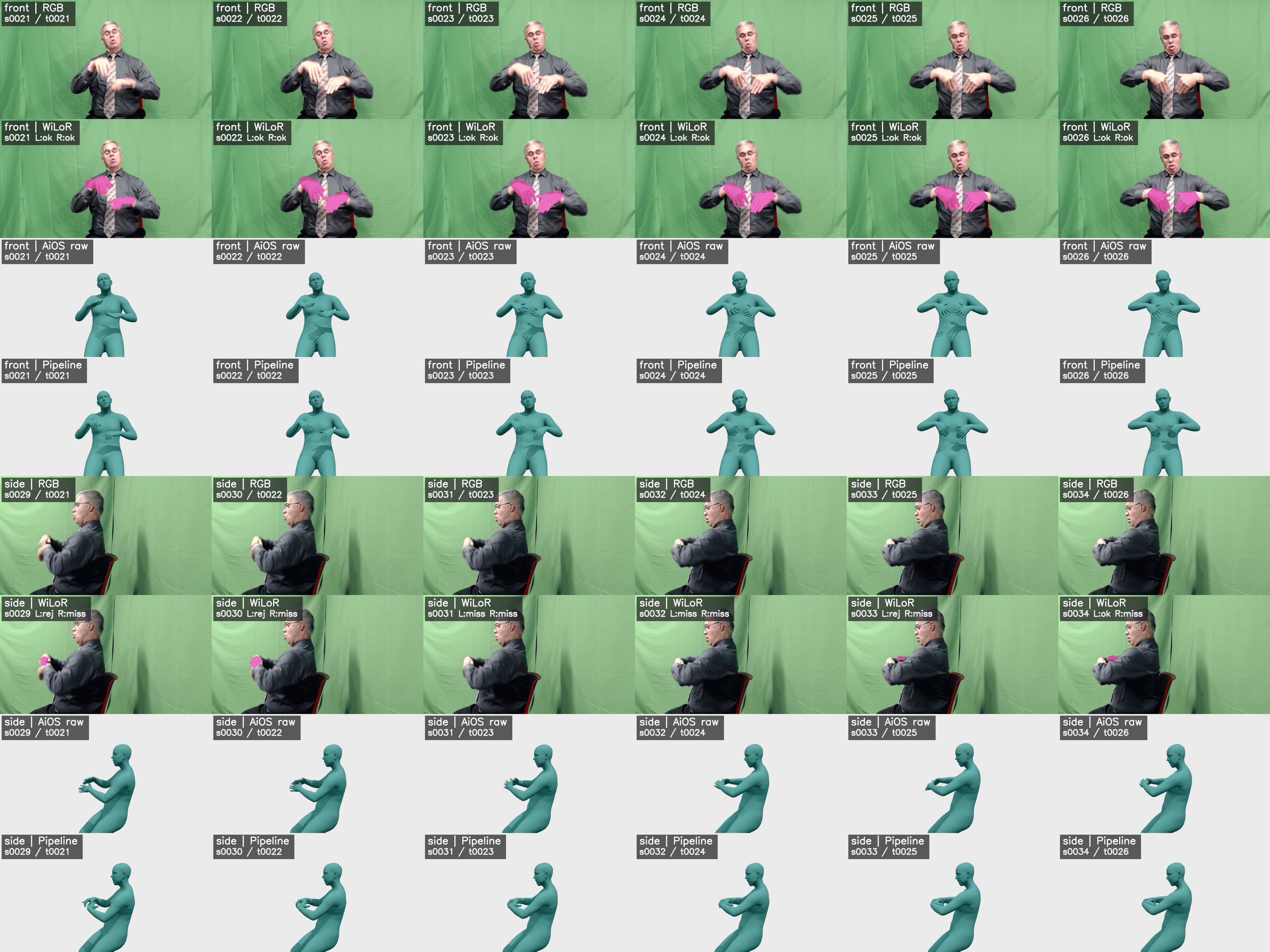}
\caption{\textbf{Qualitative example of detection absence and cross-view recovery.} Six mapped frames showing raw RGB inputs, per-view estimator outputs, and the resulting processed motion.}
\label{fig:afs6-detection-case}
\end{figure*}

\section{Sentence Boundaries and Motion-Side Non-Sign Removal}
\label{app:afs6-text-motion}

Our curation pipeline addresses temporal alignment at three distinct levels: (i) \emph{sentence-boundary segmentation}, which slices full How2Sign recordings using the official manually realigned sentence timestamps rather than the legacy How2 timing; (ii) \emph{cross-view synchronization}, which aligns frontal and side motion streams; and (iii) \emph{intra-clip non-sign removal}, which prunes internal idle intervals identically across both views using frontal kinematic cues (detailed in Appendix~\ref{app:sub_sync_and_clean}) while keeping the paired English transcript unchanged.

\paragraph{Motivation: untranscribed waiting intervals.}
Even within manually realigned sentence boundaries, raw recordings frequently contain extended pauses—such as signers waiting for the next teleprompter prompt—that have no lexical counterpart in the English text. Because hiring expert ASL annotators to manually trim every sub-sentence pause across 35k clips is prohibitive, our kinematics-driven non-sign filter provides a scalable way to compact these idle intervals. Notice that our conservative buffering rules (23-frame onset delay and 8-frame lookahead) are specifically designed to remove prolonged waiting periods while preserving brief grammatical holds and transitions.

\begin{table*}[t]
\centering
\footnotesize
\caption{\textbf{Motion-side non-sign pruning workload and per-clip source-frame deletion statistics.} Real-frame removal is frame-weighted over decoded frontal frames, while 4-frame synthetic bridges are counted separately per view. Per-clip deletion distributions are reported across all clips (\emph{All}, where $\min=0$) and restricted to pruned clips (\emph{Edited}, with $\ge 1$ removed interval).}
\label{tab:afs6-idle-removal}

\begingroup
\setlength{\tabcolsep}{2pt}
\renewcommand{\arraystretch}{1.1}

\begin{tabular*}{\textwidth}{@{\extracolsep{\fill}}llrrrrrrrrr@{}}
\toprule
 & & &
 \multicolumn{3}{c}{\textbf{Cohort Workload}} &
 \multicolumn{5}{c}{\textbf{Per-Clip Deleted Source Frames}} \\
\cmidrule(lr){4-6}
\cmidrule(lr){7-11}

Cohort &
Split &
Clips &
\shortstack{Edited\\clips (\%)} &
\shortstack{Frames\\removed (\%)} &
\shortstack{Bridge\\frames} &
\shortstack{Mean / Med\\(All)} &
\shortstack{Mean / Med\\(Edited)} &
\shortstack{Min\\(Edited)} &
\shortstack{Max\\total} &
\shortstack{Max single\\run} \\

\midrule
New & train & 26,791 &
\shortstack{16,194\\(60.45\%)} &
18.72\% & 91,176 & 32.69 / 18 & 54.08 / 37 & 11 & 2,400 & 850 \\

New & val & 1,557 &
\shortstack{966\\(62.04\%)} &
18.57\% & 5,600 & 33.45 / 21 & 53.92 / 42 & 12 & 377 & 336 \\

New & test & 2,099 &
\shortstack{1,218\\(58.03\%)} &
16.79\% & 6,672 & 29.33 / 16 & 50.54 / 36 & 12 & 457 & 436 \\

New & overall & 30,447 &
\shortstack{18,378\\(60.36\%)} &
18.58\% & 103,448 & 32.49 / 18 & 53.83 / 37 & 11 & 2,400 & 850 \\

\midrule
Reused & overall & 4,492 &
\shortstack{2,769\\(61.64\%)} &
18.89\% & 15,988 & 34.08 / 19 & 55.28 / 38 & 12 & 738 & 676 \\

\bottomrule
\end{tabular*}
\endgroup
\end{table*}

\paragraph{Dataset-wide pruning statistics.}
Table~\ref{tab:afs6-idle-removal} summarizes the non-sign pruning statistics across the 30,447 newly processed clip pairs and 4,492 reused pairs. In the primary 30,447-clip cohort, intra-clip pruning is triggered in 18,378 clips (60.36\%), removing 989,360 out of 5,325,225 decoded frontal frames (18.58\% frame-weighted) and inserting 103,448 synthetic bridge frames (4 frames per cut) per view. Because excised intervals are bridged by 4-frame transitions, we track source-frame deletion indices separately from synthetic bridges and boundary padding. In Table~\ref{tab:afs6-idle-removal}, we report both the \emph{total removed source frames} per clip and the \emph{longest contiguous removed run} per clip (evaluated across all clips as well as edited clips only) to characterize the distribution of sustained idle pauses.

\paragraph{Scope and limitations.}
While combining manually realigned sentence boundaries with kinematic pause removal substantially reduces temporal redundancy, motion-activity cues operate purely on kinematics rather than lexical semantics. Consequently, this pipeline targets untranscribed idle intervals within a clip, rather than correcting potential upstream transcript errors or performing word-level forced alignment.

\subsection{Released Resources and Statistics}
\label{app:release_scope}

The release comprises three SMPL-X motion datasets, each providing the final reconstructed motion sequence per sample (omitting raw AiOS/WiLoR predictions and per-view intermediates). Specifically, the curated outputs comprise: (i) \textbf{How2Sign-MV3D} with 34,939 sequences (30,903 train, 1,711 validation, and 2,325 test); (ii) \textbf{ASLCitizen-3D} with 83,325 sequences derived from ASLCitizen; and (iii) \textbf{MM-WLAuslan-MV3D} with 38,580 sequences derived from MM-WLAuslan. Note that these totals reflect the successfully curated motion sequences rather than the nominal sizes of the upstream video corpora.

In addition, \textbf{SGNify-MV} provides rendered frontal- and side-view videos for the 57 SGNify signs, generated from the original ground-truth meshes to support controlled multi-view reconstruction evaluation. Unlike the three motion datasets above, SGNify-MV releases the rendered multi-view videos alongside our Blender rendering scripts to allow custom viewpoint generation.

Finally, we release the GPT-5.2 How2Sign pseudo-gloss annotations, the cross-view curation pipeline, and model training scripts via our GitHub repository. To respect upstream licensing, we do not redistribute raw source videos, original SGNify meshes, or SMPL-X/MANO body model files, which remain subject to their respective access terms.

\end{document}